\RequirePackage[2020-02-02]{latexrelease}
\documentclass{clv3}

\usepackage{hyperref}
\usepackage[table]{xcolor}
\definecolor{darkblue}{rgb}{0, 0, 0.5}
\hypersetup{colorlinks=true,citecolor=darkblue, linkcolor=darkblue, urlcolor=darkblue}

\usepackage{graphicx}
\usepackage{caption}

\usepackage{hyperref}
\usepackage{amsmath, amsfonts}
\usepackage{varwidth}
\usepackage{booktabs}
\newsavebox\tmpbox
\usepackage{caption}
\usepackage{subcaption}

\usepackage{makecell}
\usepackage{tabularx}
\usepackage{lipsum}
\usepackage{mwe}
\usepackage{ragged2e}
\usepackage{dirtytalk}
\usepackage{arydshln}
\usepackage{multirow}

\usepackage{algorithm} 
\usepackage{algpseudocode}

\usepackage{microtype}

\usepackage{booktabs}
\usepackage{listings}
\usepackage{xstring}
\usepackage{soul}
\usepackage[table]{xcolor}
\lstdefinestyle{odin-style}{%
  language=odin,
  basicstyle=\ttfamily,
  xleftmargin=0pt,
  xrightmargin=0pt,
  frame=tb,
  rulecolor=\color{black},
  extendedchars=true,
  showstringspaces=false,
  showspaces=false,
  tabsize=2,
  breaklines=true,
  showtabs=false,
  escapeinside={(*@}{@*)}
}

\lstnewenvironment{odinrule}[1]{%
 \lstset{style=odin-style}}{%
\captionof{lstlisting}{#1}
}

\algrenewcommand\algorithmicrequire{\textbf{Input:}}
\algrenewcommand\algorithmicensure{\textbf{Output:}}

\issue{1}{1}{2022}

\runningtitle{OLGA}

\runningauthor{Kumar et al.}

\begin{document}

\title{OLGA : An Ontology and LSTM-based \\approach
for generating Arithmetic Word Problems (AWPs) of transfer type}

\author{Suresh Kumar\thanks{Ph.D. Scholar, SSB317, Dept. of CSE, IIT Madras, Chennai, India, E-mail: cs18d007@cse.iitm.ac.in}}
\affil{Indian Institute of Technology Madras}

\author{P Sreenivasa Kumar\thanks{Professor, SSB215, Dept. of CSE, IIT Madras, Chennai, India, E-mail: psk@cse.iitm.ac.in}}
\affil{Indian Institute of Technology Madras}

\maketitle

\begin{abstract}
Machine generation of Arithmetic Word Problems (AWPs) is challenging as they express quantities and mathematical relationships and need to be consistent. ML-solvers require a large annotated training set of consistent problems with language variations. Exploiting domain-knowledge is needed for consistency checking whereas LSTM-based approaches are good for producing text with language variations. Combining these we propose a system, OLGA, to generate consistent word problems of TC (Transfer-Case) type, involving object transfers among agents. Though we provide a dataset of consistent 2-agent TC-problems for training, only about 36\% of the outputs of an LSTM-based generator are found consistent. We use an extension of TC-Ontology, proposed by us previously, to determine the consistency of problems. Among the remaining 64\%, about 40\% have minor errors which we repair using the same ontology. To check consistency and for the repair process, we construct an instance-specific representation (ABox) of an auto-generated problem. We use a sentence classifier and BERT models for this task. The training set for these LMs is problem-texts where sentence-parts are annotated with ontology class-names. As three-agent problems are longer, the percentage of consistent problems generated by an LSTM-based approach drops further. Hence, we propose an ontology-based method that extends consistent 2-agent problems into consistent 3-agent problems. Overall, our approach generates a large number of consistent TC-type AWPs involving 2 or 3 agents. As ABox has all the information of a problem, any annotations can also be generated. Adopting the proposed approach to generate other types of AWPs is interesting future work.
\end{abstract}

\section{Introduction}
\begin{figure}
    \centering
    \includegraphics[width=0.9\textwidth]{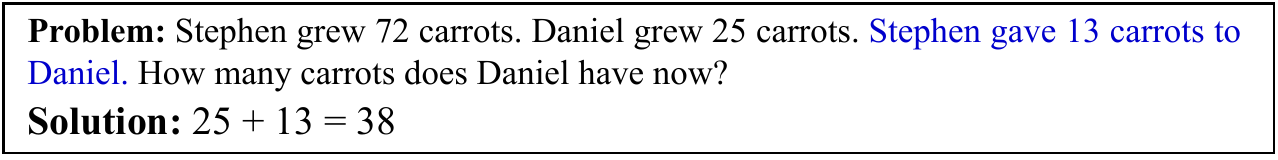}
    \caption{Example TC-type AWP}
    \label{fig1}
\end{figure}

Although the Natural Language Generation (NLG) systems generate reasonably good quality text, some of the NLG outputs do require repairs. If NLG systems are deployed to generate mathematical text, there is a need to check the consistency of the generated output. Generating domain-specific and quality text has many applications like providing more training data for tasks such as language translation, chatbots for domain-specific communication, building pedagogical tools, etc.\\
Arithmetic Word Problems (AWPs) are elementary math problems where the relevant information about the problem is present in natural language text rather than equations or notations. Transfer Case (TC) word problems are specific types of AWPs that involve transfers of objects among two or three persons (called Agents in the rest of the paper). A typical TC-AWP consists of one or more clauses/sentences, each of which belongs to one of the following sentence categories: before-transfer (BT), transfer (TR), after-transfer (AT), and question (QS). We provide more details about these sentence categories in Section~\ref{sec:tconto}. Figure~\ref{fig1} presents an example of TC-AWP. Here, the first two sentences are of BT-type, the third sentence is of TR-type, and the last sentence is of QS-type. The given TC-AWP does not involve any AT-type sentence.
The TC-AWP text typically involves agent names, quantities, quantity types, possibly some hints on how to approach the solution, etc. Humans can easily solve word problems (WPs), as they have the required domain knowledge. However, automatically solving these problems is challenging for AI systems, as it requires programmatically-modeled domain knowledge and a robust system that can "understand" the natural language variations. \cite{TC-AWP-Solver} showed how the domain ontology is very helpful for the task of solving and also detecting inconsistencies in TC-AWPs. We believe that domain knowledge would be equally helpful in \textit{generating} consistent TC-AWPs. \\
The popular English AWP datasets (which contain transfer cases) are AllArith \cite{unitdep}, ALG514 \cite{learningtoautosolve}, MAWPS \cite{mawps}, and Dolphin-S \cite{dolphin};
they are either collected from online websites such as \texttt{answers.yahoo.com}, \texttt{math-aids.com}, \texttt{k5learning.com}, etc. or they are curated from other smaller AWP datasets such as IL \cite{exptree}, AI2 \cite{AI2}, etc. The AWP datasets contain limited types of problems (example: TC-AWPs, Age AWPs, part-whole AWPs, etc.) and a relatively small number of examples in these categories. For example, the proportion of transfer cases in AllArith and Dolphin-S is less than 40\% and 10\%, respectively. The proportion of Age AWPs is even less than 1\% in these datasets. Therefore, enriching the AWP datasets by generating more word problems would provide a supportive ground for ML/DL based AWP solver systems. So far, limited attempts have been made for word problem generation.\\
% For the word problem generation task, \textbf{we choose to study the TC-AWP domain for the following reasons:} (1) by solving the case of word problem generation; we aim to provide a large set of consistent problems that the AI community can use to test various solution approaches (2) the popular AWP datasets (like AllArith) contain the highest percentage of TC word problems as compared to the other types of word problems present in these datasets, therefore, enabling us to develop a learning-based module as a part of the solution (3) The availability of the TC-Ontology, which was recently proposed by us while developing a TC-AWP solver system \cite{TC-AWP-Solver}. We reuse the TC-Ontology after required modifications.\\
We consider a word problem \textit{consistent} when all the information in the problem text is consistent with the assumptions for the domain. Although we train the LSTM model on consistent TC-AWPs, the model generates the word problem text of three kinds: (a) consistent TC-AWPs (like very similar to those which were given during the training phase), (b) partially consistent TC-AWPs (those TC-AWPs in which one or two elements are generated wrongly) - but they are repairable, (c) inconsistent TC-AWPs \textit{or} unrepairable cases - the word problem text which is beyond repairable as it lost the essential structure during the generation.
The AWP example mentioned in Figure~\ref{fig1} is of consistent type. The examples of other two categories are discussed later. In the example given in Figure~\ref{fig1}, the third sentence of the problem text is consistent with the first two sentences, as the information provided in the first two sentences is sufficient to carry out the task mentioned in the third sentence. Also, it is possible to answer the question posed in the fourth sentence with the information available in the first three sentences. Moreover, the \textit{transfer} (third sentence) mentioned in the problem text is possible, as Stephen has more quantities than what is required in the \textit{transfer}. Therefore, the TR-type sentence is consistent with the first two sentences.
In the TC-AWP subdomain, for a WP to be consistent, we consider the following factors: the agents, quantities, the order of the sentences, what is being asked in the question, etc. The existing AWP datasets contain only consistent problems. However, during the WP generation process, if a sentence-part (maybe agent-name, quantity-type, etc.) is generated wrongly, it makes the generated problem inconsistent. Additionally, it's challenging to generate consistent TC-AWPs of longer length. We note that, with an increase in the number of sentences, the chances of generating consistent WPs reduce, leading to more inconsistent cases in the generated examples. We verify this claim by including related experiments in Section~\ref{sec:experiment}.\\
Ontologies are beyond thesauri, lists, and taxonomies, as they provide vocabulary (formal description of definitions of abstract classes and their relationships) and axioms to standardize the information in a domain of interest. "Formal" means that vocabulary definitions are based on a logical framework (such as Description Logic). This enables domain representations to be machine-processable, eventually allowing reasoning, inferring new knowledge, and automatic inconsistency detection (in the semantic model and its extended application). We focus on OWL ontologies that are founded on Description Logics (DLs). 
% The authors \cite{dlfornlp} showed how encoded domain knowledge (using DLs) about syntactic, semantic, and pragmatic elements can drive the natural language generation process. However, in our work,
We use domain knowledge in the followings ways: checking the consistency of the generated WPs, repairing the partially consistent WPs, and generating multiple-transfer-WPs from single-transfer-WPs.\\
For text generation, RNN extensions that preserve the long-term memory of tokens, such as LSTMs or GRUs, have been found to perform well in practise \cite{prasad,gmail}. Typical applications of text generation systems include the generation of weather reports, chatbots, image captioning, etc. We analyze that minor errors during text generation are acceptable in some domains, for example - producing weather reports. Note that here we talk about the "generation error in weather report" not the error in weather prediction. Also, some applications require generating the text of short length, for example - chatbots, image captioning, etc. Text generation becomes challenging if we need to generate consistent mathematical text. 
% For more concrete analysis, consider the following example WP: \textit{Stephen has 5 books and 2 pens. Daniel has 12 books. Stephen gave 2 books to Daniel. How many books does Daniel have now?}. In order to successfully generate such example, system needs to generate word problem sentences that are consistent with each other.
Since language variations in the generated text is important in a good dataset, we use LSTMs for generation and subsequently check the consistency of the generated problems.
To the best of our knowledge, none have proposed a system that automatically identifies the consistent cases from the generated examples and also repairs the partially consistent cases. Also, the existing datasets do not contain multiple transfer AWPs.\\
\textbf{Overview of the proposed system \& key contributions:} \\
% \hspace{-3.5mm}\textit{Research Problem:} Propose a system that leverages an LSTM-based module to generate transfer-type word problems and, over the generated examples, uses ontology-based modeling to identify the consistent cases; and also repairs the partially-consistent word problems and makes them consistent.\\
We present an ontology and LSTM-based combined approach for TC-AWP generation. Note that purely ontology-based TC-AWP generation, even though it achieves consistency, is deficient in language variations (expected in such datasets); hence it can not be adopted as a preferred method. We gather TC-AWPs from popular AWP datasets and name the dataset AllArith-Tr. First, we train the LSTM-based text generation model using AllArith-Tr and generate many TC-AWPs. We analyze the generated examples and identify three significant categories: consistent TC-AWPs, partially consistent TC-AWPs (or repairable cases), and inconsistent TC-AWPs (unrepairable cases). Our key contribution is the ontology-based approach that identifies the consistent cases from the generated examples and makes the partially-consistent-WPs consistent. The domain knowledge required in the current modeling is encoded using Ontology T-Box and a set of Semantic Web Rule Language (SWRL) rules. We capture the important information from word problem text using Ontology A-Box. To automate the A-Box extraction, we annotate the word problem sentence-parts with class-names from the domain ontology and train a BERT language model that learns to produce the instantial information for the ontology classes. The consolidated knowledge (T-Box plus A-Box) helps check the consistency of the generated examples and make the partially-consistent word problems consistent. The complete architecture diagram of the proposed system is given in Figure~\ref{fig4}. The key contributions of work are:
\begin{enumerate}
     \item \justifying Incorporating domain knowledge to solve AI challenges, such as AWP-solving and AWP-generation, is an exciting topic in Natural Language Understanding (NLU). We show a way to leverage learning \& knowledge-based technologies to solve the AWP generation challenges effectively.
     \item Ours is the first attempt in the direction of detecting \& fixing the inconsistencies in the word problem text. 
    %  \item Since our primary goal is to utilize domain knowledge, we present extended TC-Ontology (we proposed it for solving TC-AWPs) to formally represent concepts, relationships, and axioms for the TC domain. We model the TC domain's knowledge using appropriate axioms and SWRL rules. Also, after the required modifications, the domain ontology can be reused in other applications, such as generating explanations for AWP solutions.
    \item We extend the TC-Ontology, which was proposed in our earlier work, and capture the knowledge required for consistency-checking and repair process.
     \item Further, the proposed system can generate TC-AWPs that involve multiple transfers. Existing AWP datasets do not contain multiple-transfer WPs.
\end{enumerate}
Sections~\ref{sec:prelim} \&~\ref{sec:relatedwork} discuss the preliminaries and related work, respectively. Section~\ref{sec:olga} presents the proposed approach OLGA. Section~\ref{sec:multitransfer} discusses the extension of the OLGA system for generating multiple transfer TC-AWPs. Section~\ref{sec:experiment} provides implementation details and results analysis. At last, in Section~\ref{sec:conclusion}, we conclude and discuss future directions.
%At last, we discuss the conclusion of the work and provide future directions in Section~\ref{sec:conclusion}.
\section{Preliminaries - Text Generation, Ontology}
\label{sec:prelim}
This section discusses text representation techniques, the process of generating text, popular text generation models, and an introduction to ontologies.
\subsection{Text : representation \& generation}
Many NLP applications rely on utilizing suitable sentence representations. The non-distributive representations suffer from sparsity and dimensionality issues, whereas the distributive representation of words in vector space enables text generation models to perform better \cite{touseef}. The popular techniques to learn distributive representations are Word2Vec \cite{word2vec}, GloVe \cite{glove}, FastText \cite{fasttext}, etc.\\
A text generation model learns a probabilistic distribution that represents the training data. Let $P(w_{t}|w_{t-1},w_{t-2},...,w_{t-n})$ be the probability distribution of a token at time-step $t$, given a sequence of previous $n$ tokens. In $t+1^{th}$ time-step, the probability distribution tries to represent $w_{t+1}$ given $w_{t},w_{t-1},...,w_{t-n}$. The process is continued until a convergence condition is met, and the tokens are then organised in a sequential order to form the generated text. Text generation models encode features and rules from the training data/text during the training phase. The popular text generation models are RNNs \cite{rnnoriginal}, LSTMs \cite{lstmoriginal}, GRUs \cite{grufornl}, Variational Auto-Encoders (VAEs) \cite{vaefornl}, Generative Adversarial Networks (GANs) \cite{ganfornl}, etc. However, picking a suitable model depends on the domain and the objective at hand.
\subsection{Ontology}
Ontology is a formal knowledge modeling framework that explicitly represents classes (also called concepts), designated properties (also called roles/slots) describing various features and attributes of the concept, and restrictions on roles, in a domain of discourse. Ontologies have a wide range of applications: semantic web, medicine, geospatial reasoning, program analysis, etc. Resource Description Framework-Schema (RDFS) \cite{RDFS} and Web Ontology Language (OWL) \cite{owl} are two widely used frameworks to process ontologies computationally and they differ in their expressive capabilities. Resource Description Framework (RDF) \cite{rdf} is a data modeling standard that allows effective information exchange across the web with reasoning enabled. It is used to build RDFS and OWL technologies. In the following, we discuss essential technologies required to understand ontologies.\\
\textbf{\textit{a) Resource Description Framework (RDF):}} RDF is a resource description framework, enabling the encoding, reusing, and exchanging of structured metadata. A resource can be a document, people, physical object, or abstract concept. RDF represents a domain's resources by making statements about them. The structure of an RDF statement consists of three parts (<subject \textit{predicate} object> or <S P O>), called a triplet, where a Unique Resource Identifier (URI) uniquely identifies each part. For example, the RDF triple representation of an English sentence "Person name is Stephen" would be as follows: <http://xmlns.com/foaf/spec/\#term\_Person  \textit{http://xmlns.com/foaf/spec/\#term\_name} http://example.name/\#Stephen>; where S, P, and O elements are the URIs representing a class \textit{Person}, a property \textit{Person-Name}, and the person \textit{Stephen}, respectively.\\
A URI is a character string with two elements: \textit{namespace} and \textit{local-name}, and it has a global scope. The \textit{namespace} denotes the URI's fixed prefix-component, whereas the \textit{local-name} denotes the URI's varying suffix-component. The \textit{namespace} for a domain's linked URIs is unique. RDF allows giving each \textit{namespace} an abbreviation for ease of use. Representatively, each URI takes the form \textit{abbreviation:local-name}. For example, by using the abbreviations \textit{foaf} for http://xmlns.com/foaf/spec/ and \textit{ex} for http://example.name/  we can compactly write the above triple as <foaf:\#term\_Person \textit{foaf:\#term\_name} \textit{ex}:Stephen>. W3C defined some URIs for uniform usage and made them available under the \textit{namespace} http://www.w3.org/1999/02/22-rdf-syntax-ns\# (abbreviation: \textit{rdf}). However, while modeling the domain-specific requirements, one can define the custom namespace and appropriate abbreviation.\\
%Note that RDF, on its own, does not provide a way to structures its resources. \\
\textbf{\textit{b) Resource Description Framework-Schema (RDFS):}} RDFS is a vocabulary description language and provides a way to add schema to RDF. RDF vocabulary is a set of classes with specific properties that use the RDF data model to provide essential elements to model a domain. It also specifies axioms or entailment rules to infer new triples. On a summary note, RDFS enables modeling a domain-specific vocabulary, also known as minimal ontology, and can be used to make \& infer domain statements.\\
\textbf{\textit{c) Web Ontology Language (OWL):}} \textit{OWL} is a Semantic Web language family designed to formally model complex \& rich knowledge of a domain of interest. Since it is a computational logic-based language, computer programs can exploit the knowledge expressed in OWL. OWL extends RDFS and allows adding more constructs for describing classes and properties, e.g., cardinality constraints, modeling disjoint classes, etc. W3C approved three variants of the OWL differing in increasing expressiveness, namely OWL-Lite, OWL-DL, and OWL-Full. Since the OWL-DL variant provides maximum expressiveness and retains computational completeness and decidability, we use OWL-DL ontology in our modeling. OWL-DL ontologies are founded on the Description Logic (DL) foundations; hence, this correspondence gives it its name. The Description Logics (DLs) \citep{dlbook} are decidable first-order logic (FOL) fragments. OWL-DL is based on the $\mathcal{SHOIN^{(D)}}$ description logic \cite{ontology-handbook}. An OWL-DL ontology $\mathcal{O}$ can be understood as a pair ($T$, $A$), where $T$ represents Terminological-Box (TBox) and $A$ represents Assertional-Box (ABox). By utilizing vocabulary terminology, TBox captures the definitions of the concepts and properties, whereas ABox is used to provide the membership assertions - either as concepts or as properties and is the place where the details of a given concrete situation in the domain are presented.\\
\textbf{\textit{d) Semantic Web Rule Language (SWRL):}} \cite{swrl} developed a rule language for the Semantic Web, abbreviated as SWRL, to extend the OWL axioms by including the Horn-like rules. Even though OWL brings sufficient expressive power to the Semantic Web, it has expressive limitations, particularly the assertions that need to utilize more than one property. SWRL helps overcome these limitations. Moreover, DL-safe SWRL provides feasible reasoning for OWL-DL setting \cite{dlsafeswrl}. A DL-safe SWRL rule takes form $a_1 \wedge a_2 \wedge ... \wedge a_k \rightarrow a_{k+1} \wedge a_{k+2} \wedge ... \wedge a_{n}$, where $a_i$ represents an atomic unit. Here, every atom is either a Class[C(a)] or a property[P(b, c)], where a, b, and c represent either individuals or variables.
\section{Related Work}
\label{sec:relatedwork}
Our survey on related work primarily discusses two aspects: a) deep learning models for text generation and b) the approaches that focus on word problem generation.
%We survey the available methods for \textit{text generation} and approaches that focus on \textit{word problem generation}, which are relevant to our work, and a summary of the approaches is given below.
\subsection{Deep learning models for text generation}
In this section we discuss the deep learning models available for text generation. We mainly focus on the architectural difference of these models and their pros and cons.\\
\textbf{Recurrent Neural Networks (RNNs):} are the most powerful algorithm for sequence modeling task \cite{touseef}. RNNs have internal memory, which allows them to remember both prior and current inputs, making sequence modeling tasks more accurate \cite{touseef}. The output at any timestamp depends not only on the current input but also on the output generated at previous timestamps, making it ideal for tasks such as text generation, machine translation, sentiment analysis, etc. Training RNNs is difficult due to the exploding/vanishing gradient problems \cite{fundaOfRNNLSTM}. The important equations that are used in the RNNs are given below.
% \begin{equation*}
%     h_{t} = \sigma (W_{xh}^{T}x_{t} + W_{hh}^{T}h_{t-1})\\
%     \hat{y_{t}} = W_{hy}^{T}h_{t}
% \end{equation*}
% \begin{equation}
%     \hat{y_{t}} = W_{hy}^{T}h_{t}
% \end{equation}
\begin{align} 
h_{t} = \sigma (W_{xh}^{T}x_{t} + W_{hh}^{T}h_{t-1})  \\
 \hat{y_{t}} = W_{hy}^{T}h_{t}
\end{align}
Here, equation (1) models the hidden state, whereas equation (2) represents the generated output. The following are the descriptions of the variables used in the above equations- $\sigma$: activation function, $x_t$: input vector at $t^{th}$ timestamp, $\hat{y_{t}}$: output vector, [$W_{xh}$, $W_{hh}$, $W_{hy}$]: are the various weight matrices used in the network, $h_{t-1}$: represents the previous hidden state. Biases are ignored. Note that the high dimensional hidden states and the recurrent connection allow RNNs to capture the information related to sequence modeling \cite{fundaOfRNNLSTM}. 
\begin{figure}[t]
    \centering
    \includegraphics[scale=0.6]{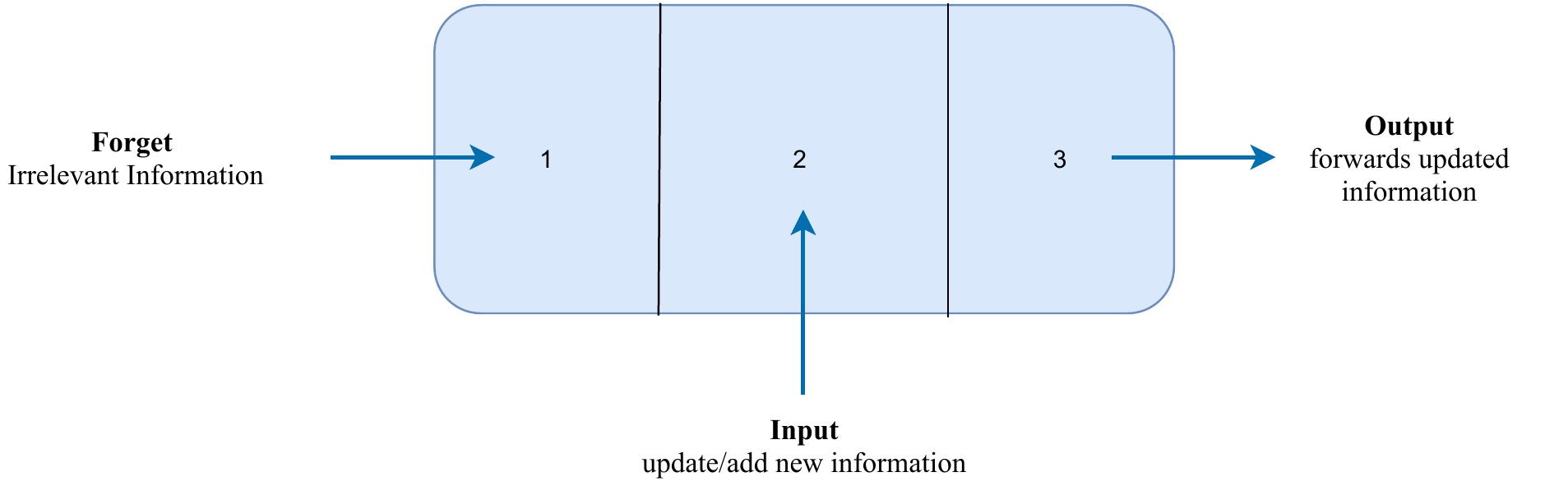}
    \caption{High level overview of an LSTM cell}
    \label{fig2}
\end{figure}\\
\textbf{Long Short Term Memory (LSTM) and Gated Recurrent Unit (GRU):} LSTM/GRU networks inherit the same basic RNN architecture; however, there are no hidden states. The information is added or removed through structures called gates. These structures are "computational blocks" that control the flow of information. LSTMs/GRUs learn to keep relevant information to make predictions and forget all the non-relevant information. An \textbf{LSTM} unit uses three gates to process the information: Input Gate, Forget Gate, and Output Gate. Figure~\ref{fig2} shows a high-level description of these three gates. The first portion represents \textit{Forget Gate}, and it determines whether the information from the previous timestamp should be remembered or is irrelevant and can be discarded. The second part of the cell, known as \textit{Input Gate}, is responsible for learning new information from the current input. The third portion, called \textit{Output Gate}, forwards the updated information from the current timestamp to the next. The following equations represent the mathematical modeling of these three gates.
\vspace{-1mm}
\begin{equation}
   o_{t} = \sigma(W_{o}h_{t-1} + U_{o}x_{t} + b_{o}) \text{; } i_{t} = \sigma(W_{i}h_{t-1} + U_{i}x_{t} + b_{i}) \text{; } f_{t} = \sigma(W_{f}h_{t-1} + U_{f}x_{t} + b_{f})
\end{equation}
% \begin{equation}
%     i_{t} = \sigma(W_{i}h_{t-1} + U_{i}x_{t} + b_{i})
% \end{equation}
% \begin{equation}
%     f_{t} = \sigma(W_{f}h_{t-1} + U_{f}x_{t} + b_{f})
% \end{equation}
In Eq. 3, we present the equations of all three LSTM gates at once, separated by a semicolon. The notations $o$, $i$, and $f$ represent output, input, and forget gate, respectively; $\sigma$: activation function; $W$, $U$, and $b$ are the matrices representing weights used in the network; $x_t$: current input; $h_{t-1}$: output of the previous timestamp.\\
A \textbf{GRU} unit uses two gates to process the information: Update and Reset. The \textit{update} gate computes the amount of previous information that needs to pass along to the next cell state.
The \textit{reset} gate is responsible for determining how much of the past information is irrelevant and can be ignored. GRUs execute faster as they involve fewer training parameters, whereas LSTMs are more accurate on datasets containing longer sentences.\\
\textbf{Variational Auto-Encoders (VAEs):} For the text generation task, VAE networks employ an encoder to convert data/text into latent variables, and then the decoder uses these latent variables to generate the data/text. Let $x$ be the input for encoding operation, then the VAE model produces latent representation $P_{\theta}(y|x)$, where $\theta$ represents the parameters of the encoding operation. In contrast, the decoder tries to find the probability distribution $q_{\phi}(x|y)$ of data on given latent distribution $P$, where $\phi$ represents the parameters of the decoding operation. The VAEs have emerged as a popular method for unsupervised learning of complex distributions. However, the applications of VAEs for the text generation task are limited, as the text data is discrete in nature \cite{touseef}. Also, the VAE-based text generation approaches face difficulty while generating the longer text sequences \cite{touseef}.
\\
\textbf{Generative Adversarial Networks (GANs):}
GANs are the popular deep learning algorithms that adopt an adversarial training strategy. GANs are composed of two models- Generator and Discriminator.
The generator model generates data samples intending to keep the generated data samples very close to the true data.
The objective of the discriminative model is to classify the generated data samples into real (training data) or fake (the generated ones).
GANs have shown promising results for the image generation tasks (Conditional GANs, Pix2Pix GANs, etc.). However, training GANs for text-generation tasks is challenging because of the non-differentiable property of discrete symbols and text data is discrete \cite{survey2}. Additionally, they need enormous data for training and are slow while generating the text.
\subsection{Approaches for word problem generation}
So far, only limited attempts have been made to solve the challenge of word problem generation. \cite{automatic} is one of the earliest work in this direction. The work adopts an NLG architecture proposed in \cite{building} and leverages the concept of Frame Semantics to investigate distance-rate-time WPs. However, the details related to the assessment of the system (such as accuracy, benchmarks, etc.) are not provided, which makes it hard to assess the system. \cite{theme} proposes an approach that generates new WPs by re-writing the theme of the existing WPs. However, the generated WPs involve a similar underlying story and a same set of equations. For example - the WP "Stephen has 12 books. He gave 8 books to Daniel. How many books does Stephen have now?" can be used to generate a new WP with a different theme, which is as follows: "Elena had 12 pens. She gave 8 pens to Mike. How many pens does Elena have now?". Note that, in this approach, the generated WP involves same quantities and equations. \cite{personalized} focuses on generating personalized word problems from general specifications using answer-set programming (ASP). The specifications include the teacher and student requirements. The system makes use of the logical encoding of the specifications, synthesizes a WP narrative and its mathematical model as a labeled logical plot graph, and realizes the WP in natural language.\\
\\
\cite{generating} presents an ontology based approach for WP generation. The system leverages an existing ontology verbaliser to render the logical statements expressed in OWL as English sentences. The approach first takes the individuals available in the ontology and then uses the SWAT tool \cite{swat} to convert the lexical statements into English sentences. For example, the verbaliser converts a property \textit{hasType} to a natural language statement "is a kind of". The approach then groups together the generated sentences and forms a WP text. The approach also introduces the factors that control the difficulty level of WPs, such as - order of the statements, readability, inclusion of extraneous information, conceptual difficulty, etc. The approach focuses on only those OWL statements which are class assertions or property assertions, and verbalizes them. The work does not include the axiomatic OWL statements in their modeling, and also, the reasoning capability of the OWL framework is not being utilized. In contrast, our approach includes axiomatic knowledge and utilizes the reasoning facilities. Moreover, the author states that the proposed approach is generic and requires further testing.\\
\citep{towards} proposes a deep learning based approach for generating word problems from equations and topics. The system uses two RNN encoders, where the first encoder maps equations to hidden vectors, and the second encoder maps topics to word representations. Then, a fusion mechanism incorporates the information of both equations and topics and feeds it to a decoder which generates word problems. The system requires a substantial amount of annotated data (e.g., equations) for training.\\
To the best of our knowledge, none of the existing approaches provide the feature of checking the consistency of the generated word problems. Moreover, we show how to generate variations of existing word problems (e.g., generating multiple transfer cases from single-transfer TC-AWPs) using domain knowledge and reasoning.
\section{Proposed Approach: OLGA}
\label{sec:olga}
\begin{figure}[t]
    \centering
    \includegraphics[scale=0.6]{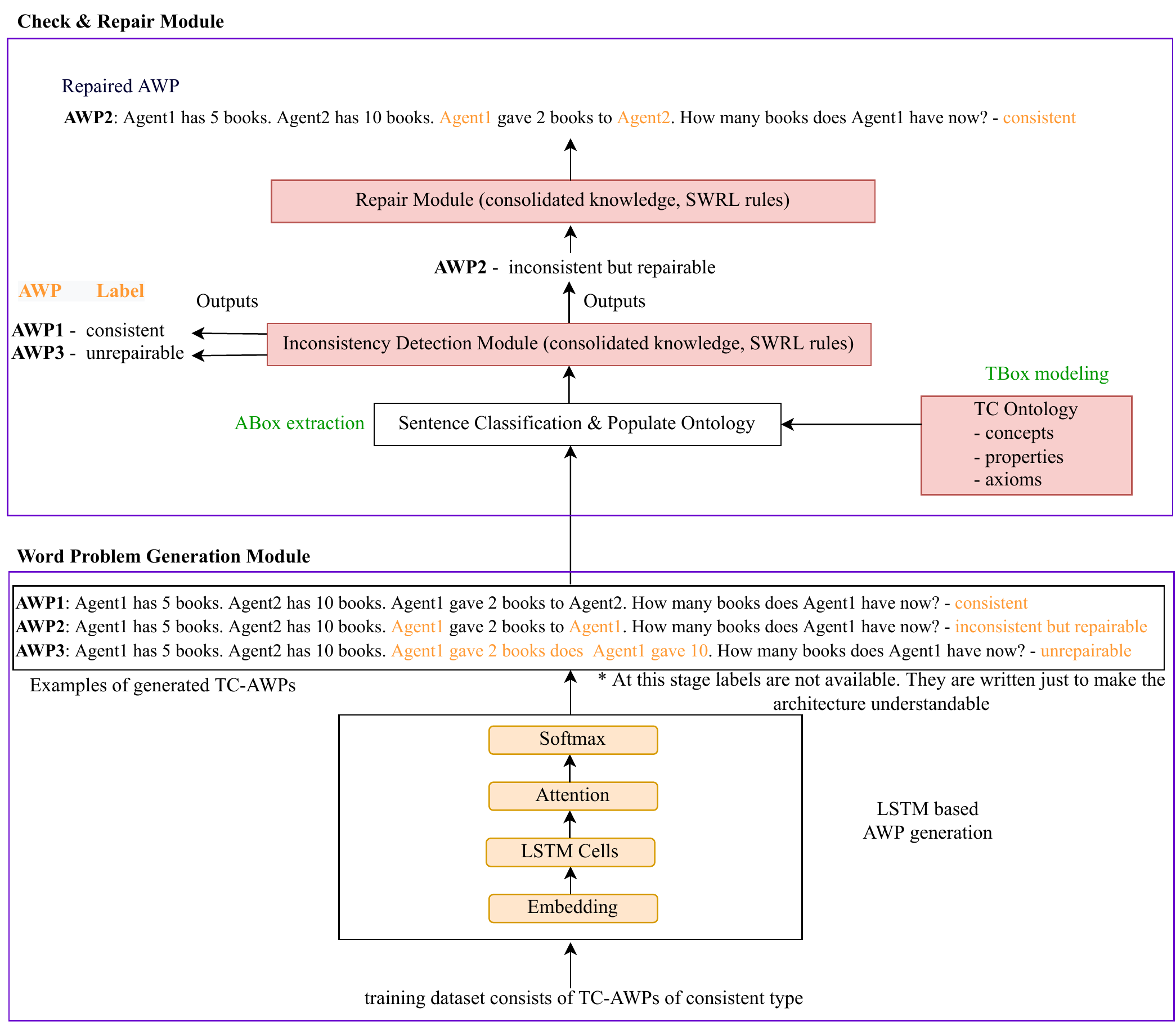}
    \caption{OLGA system architecture along with examples}
    \label{fig4}
\end{figure}

Primarily, the proposed system has two modules: "word problem generation" and "check \& repair". Figure~\ref{fig4} presents the proposed system architecture. The generation module is an LSTM model that learns to generate TC-AWP text. In contrast, the "check \& repair" module processes the LSTM-generated TC-AWPs to identify the consistent cases and repairs the partially-consistent cases to make them consistent. The "check \& repair" module consists of a domain ontology (TC-Ontology) and a set of SWRL rules. We proposed TC-Ontology while developing the TC-AWP solver system in an earlier work \cite{TC-AWP-Solver}. In the current work, we extend TC-Ontology with the constructs required for identification \& fixing of the inconsistencies and use it in the "check \& repair" module. With ontology-based post-processing (i.e., achieved using check \& repair module) of LSTM generated examples, our approach can achieve the following: (a) checking the consistency, (b) repairing partially consistent cases, and (c) generating annotations as per need. During the WP generation process, users can utilize the information available in the ontology (such as agents, to-agent, from-agent, etc.) and generate annotations. However, this work does not focus on generating annotations.
\\
During modeling the domain knowledge, we essentially model the TBox (important concepts, properties, axioms). ABox information depends on the TC-AWP at hand and needs to be populated in the ontology on the fly.
The ABox information that populates into the ontology differs based on the type of a sentence. TC-AWPs involve four types of sentences: before-transfer (BT), transfer (TR), after-transfer (AT), and question (QS). We introduce four ontology classes to model the TC-AWP sentences and treat each sentence in the word-problem-statement as an object/individual that belongs to one of these classes. The definitions of these classes are provided in Section~\ref{sec:vocab} that explains the vocabulary of the TC-Ontology. The sentence classification module provides sentence-type information to the system. While developing the domain ontology, we model the classes and properties in such a way that the information extracted from the word problem sentences can be appropriately populated into the ontology. For example, a BT type sentence such as "Agent1 has 5.0 books" contributes the following A-Box assertions: "Agent1" is an individual of \textit{Agent} class, quantity Q1 is an individual of \textit{TC-Quantity} class, object property assertion (<tc:Agent1 \textit{tc:hasQuant} tc:Q1>), and data property assertions (<tc:Q1 \textit{tc:quantValue} "5.0"> and <tc:Q1 \textit{tc:quantType} "book">). To automate the "populate-ontology" task, we use BERT-based LM \cite{bert}.\\
The generated TC-AWPs, if found consistent, are accepted as they are. In contrast, the inconsistent cases are checked for possible repairs. We analyze the major issues that cause inconsistency and encode the knowledge (required to convert inconsistent cases into consistent ones) using Ontology axioms and SWRL rules. The repair part uses the consolidated knowledge (TBox and ABox) and attempts to make the partially-consistent cases consistent if it lies within the scope of encoded knowledge. OLGA identifies and rejects the unrepairable cases and does not count them in the successfully generated cases. We detail the proposed approach in the following:
\subsection{LSTM module : Word Problem Generation}
\label{sec:lstm}
\begin{figure}
    \centering
    \includegraphics[width=\textwidth]{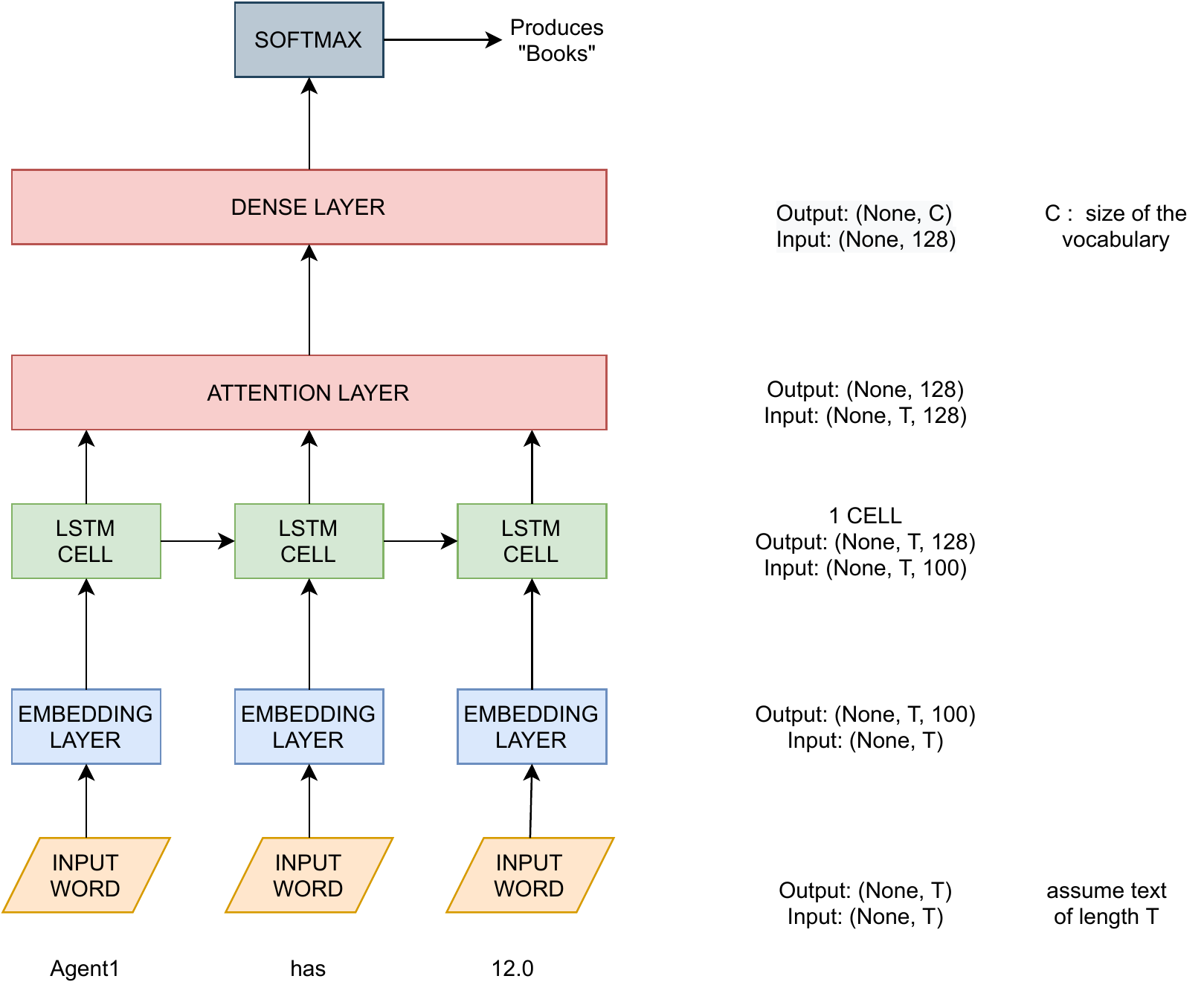}
    \caption{Architecture of the TC-AWP generation model and an example illustration}
    \label{fig3}
\end{figure}

As mentioned earlier, we focus on transfer cases (TC) from the AWP domain in this work. We choose the LSTM model for word problem generation as we primarily focus on generating consistent word problem text. For text generation, one can train the LSTM model on character-level or word-level. In this work, we adopt the word-level language model. Figure~\ref{fig3} presents the LSTM network architecture for the TC-AWP generation task. We leverage skip-embedding and attention-mechanism to improve the model quality and accelerate training. In the case of skip-embedding, the concatenated input is given to the attention layer. Note that contexts of words in natural language sentences provide additional semantic information. Therefore, the generation model includes the context information through attention-mechanism to achieves better performance. Moreover, we deliberately avoid the process of stop-word removal, as the features, such as words representing the beginning and completion of a sentence, semantic flow from the previous sentence to the following one, etc., are preserved when the model is trained with continuous words in training-text.
As explained in the example given in Figure~\ref{fig3}, the LSTM model generates the word "Books" given a sequence of words "Agent1 has 12.0". The process is continued until a complete "candidate sequence" \textit{or} "word problem" is generated. A candidate sequence is considered complete when the generation model generates a special end-of-sequence token (? in the case of word problem generation) or when the model generates a predefined maximum output sequence length.\\
The LSTM module generates grammatically stable word problem text, but not all generated WPs are consistent. The extended-TC-Ontology (part of check \& repair module), proposed in this work, takes the LSTM-generated examples for further processing and outputs only consistent TC-AWPs. Section~\ref{sec:tconto} discusses the existing TC-Ontology, as it is essential to understand Section~\ref{sec:extconto}, which presents the extended TC-Ontology (extends domain knowledge required in the current modeling). Note that Sections~\ref{sec:tconto} and ~\ref{sec:extconto} jointly explain the TBox part, whereas Section~\ref{sec:abox} explains the ABox part (essentially how it's extracted automatically) of the TC-Ontology. 
\subsection{Transfer Case (TC) Ontology - existing definitions \& constructs}
\label{sec:tconto}
We proposed TC-Ontology for TC-AWP solving \citep{TC-AWP-Solver}. 
In this work, we extend TC-Ontology to model the important concepts, relations, and other semantic information required in the current task (i.e. identifying \& fixing the inconsistencies in the LSTM-generated TC-AWPs). In the TC-AWP domain, \textit{type-of-quantity}, \textit{type-of-sentence}, \textit{direction-of-object-transfer}, etc., is the useful semantic information to model. 
In TC-Ontology, the TBox (existing and extended constructs) represents knowledge about the structure of the TC-AWP domain. In contrast, the ABox represents knowledge about a concrete situation (i.e., knowledge extracted from TC-AWP text).
\subsubsection{Vocabulary}
\label{sec:vocab}
\begin{figure}[t]
    \centering
    \includegraphics[width=\textwidth]{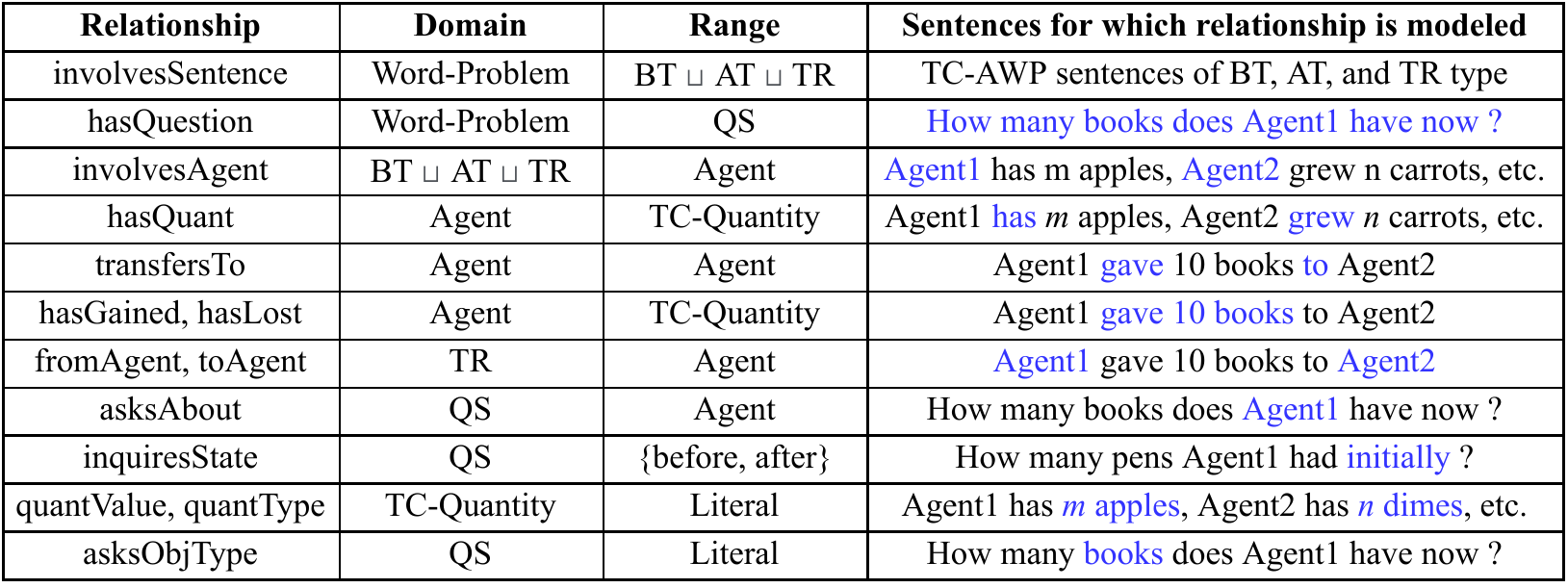}
    \caption{Important relationships from the TC-AWP domain. The \textcolor{blue}{text highlighted} in blue represents the specific part of the sentence for which the relationship is required.}
    \label{figexamplesen}
\end{figure}

Vocabulary of an ontology consists of the names (also known as terms) of the concepts/classes and relationships used to model a domain of interest. It also defines possible constraints on using these terms and reflects design choices. In the following, we discuss the vocabulary of TC-Ontology.\\
\textit{a) Concepts/Classes:} We consider each AWP as an individual belonging to the class \textit{Word-Problem}. The sentences of a word problem are categorized into individuals belonging to the following classes: \textit{BeforeTransfer (BT)}, \textit{Transfer (TR)}, \textit{AfterTransfer (AT)}, and \textit{Question (QS)}. BT class includes the TC-AWP sentences that carry the agent-quantity and associated information \textit{before} the object transfer. Similarly, AT class models the sentences that carry the post-transfer information. TR class consists of the sentences that contain object transfer information. QS class models the query-sentences that might seek the information from the followings: before-transfer facts, after-transfer facts, or transfer that is being carried out. To represent the knowledge present in a sentence, we devise the following concepts: \textit{Agent}, \textit{TC-Quantity}, \textit{PositiveQuantity}, \textit{NegativeQuantity}, etc. For example, all agents and quantities present in a sentence become individuals of the classes \textit{Agent} and \textit{TC-Quantity}, respectively.\\
\textit{b) Relationships:} In Figure~\ref{figexamplesen}, we mention the important relationships for the TC-AWP domain and mention their domain and range. A domain is a class to which the subject of an RDF statement using a given relationship/property belongs, while a range is the class of its object (value). Also, we give example sentences for which these relationships are required. For example, the \textit{hasQuant} relationship is modeled to represent a situation where an agent owns a quantity (for example- Agent1 has \textit{m} apples). The other relationships can be understood in a similar manner.
\\
\textit{c) Design Choices \& Constraints:}
Since the structures of BT and AT types of sentences are similar, the same set of relationships can be used to model these two type of sentences.
A TC-AWP might involve one or more TR-type sentences. Therefore, we model transfer as a concept and consider each specific transfer as an individual of the concept \textit{Transfer}. We make use of the \textit{predicates}: hasTR, fromAgent, toAgent, transfersTo, hasGained, and hasLost, to model a \textit{transfer}. To model the information given in a question sentence, we use the predicates: hasQuestion, asksAbout, inquiresState, and asksObjType. The domain and range descriptions of these predicates are given in Figure~\ref{figexamplesen}. In order to represent some specific domain knowledge, we apply constraints in the modeling; for example- we model \textit{TC-Quantity} as a subclass of the class \textit{PositiveQuantity}, as a negatively quantified object does not make any sense in the TC-AWP domain.
\\
\textit{d) Namespace:} The namespace of the TC-AWP domain is abbreviated as \textit{tc}. Using it, a domain sentence "Agent1 grew 10 carrots" is represented by the following triples: $<$\textit{tc:Agent1 tc:hasQuant tc:Q$>$, $<$tc:Q tc:quantValue "10"$>$, $<$tc:Q tc:quantType "carrots"}$>$.
\vspace{-1mm}
\subsubsection{Axioms}
Ontology axioms model the relationships among the concepts and set the overall theory of the domain of interest. Intuitively, axioms are logical statements that say what is true in an application domain. In the following, we mention the important axioms devised for TC-AWP domain. Here, A.01 to A.04 are concept inclusion axioms, whereas A.05 and A.06 are concept equivalence axioms.\\
A.01: ~\ensuremath{\exists}~\textit{hasQuant}.~\textit{TC-Quantity}~\ensuremath{\sqsubseteq}~\textit{Agent}~  \hspace{.45cm} (Anyone who owns a TC quantity is an agent)\\
A.02: \textit{TC-Quantity}~\ensuremath{\sqsubseteq}~\textit{PositiveQuantity} \hspace{0.72cm} (Every TC quantity is a positive quantity)\\
A.03: \textit{MinuendQuantity}~\ensuremath{\sqsubseteq}~\textit{TC-Quantity}~\\
A.04: \textit{SubtrahendQuantity}~\ensuremath{\sqsubseteq}~\textit{TC-Quantity}~\\
A.05: \textit{MinuendQuantity}~\ensuremath{\equiv}~\textit{TC-Quantity}~\ensuremath{\sqcap}~\ensuremath{\exists}~\textit{isOwnedBy}~.\textit{Agent}\\
A.06: \textit{SubtrahendQuantity}~\ensuremath{\equiv}~\textit{TC-Quantity}~\ensuremath{\sqcap}~\ensuremath{\exists}~\textit{isGainedBy}~.\textit{Agent} ~\ensuremath{\sqcap}~\ensuremath{\exists}~\textit{isLostBy}~.\textit{Agent}\\
(A.06 expresses ``Subtrahend quantity is a TC quantity that is gained by an agent and lost by an agent'')
\subsection{Transfer Case (TC) Ontology - extended definitions \& constructs}
\label{sec:extconto}
\begin{figure}[t]
    \centering
    \includegraphics[width=\textwidth]{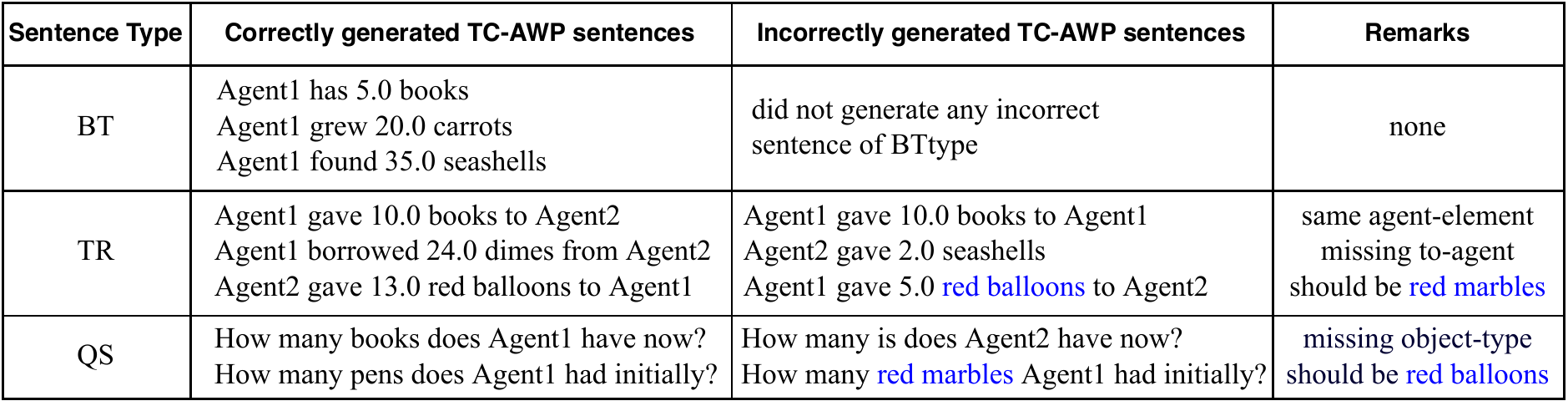}
    \caption{A category wise syntactical analysis of LSTM generated AWP sentences. The remarks mentioned directs to the wrongly generated part of the sentences}
    \label{lstmgeneratedawps}
\end{figure}

\begin{figure}[t]
    \centering
    \includegraphics[width=\textwidth]{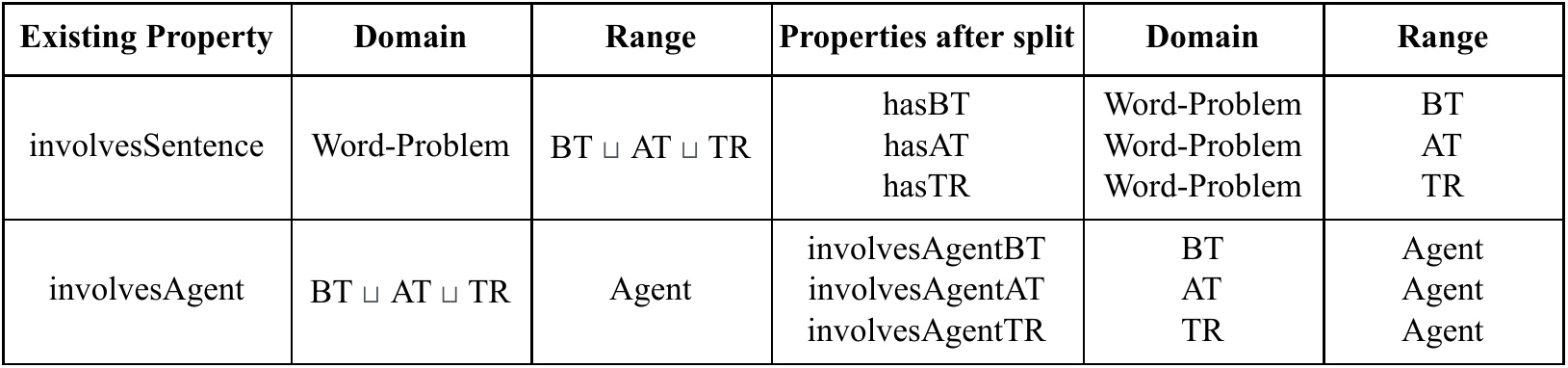}
    \caption{Property split details}
    \label{propertysplitdetails}
\end{figure}

Before we discuss the extended definitions and constructs, it is important to explain the need for the extension. In Figure~\ref{lstmgeneratedawps}, we report the correctly and incorrectly generated TC-AWP sentences. Since BT-type sentences appear at the beginning of a word problem and the LSTM model starts the generation with BT-type sentences and uses no prior information at this point, the model generates grammatically stable BT-type sentences. Also, no other sentences (such as - TR, AT, QS-type) are available at this stage to check the consistency. Therefore, all the BT-type sentences are taken as consistent. Hence, in Figure~\ref{lstmgeneratedawps}, we do not mention any wrongly generated BT-type sentence. While generating the TR-type sentences, the model uses the BT-type sentences as prior information and might generate some elements wrongly. In Figure~\ref{lstmgeneratedawps}, we mention some wrongly generated TR-type sentences. Similarly, while generating AT and QS-type sentences, the model might generate some elements wrongly.
The analysis of the "incorrectly generated WP sentences" recognizes two major scenarios: 1) repairable case: the generated sentence is inconsistent, as an element of it is generated wrongly and 2) unrepairable case: the generated sentence is inconsistent, as it lost the essential structure (as compared to the structures of TC-AWP sentences) during the generation process. We use the extended modeling in identifying and fixing the repairable cases. Also, our system identifies the unrepairable cases. However, fixing these cases is deemed not really worth the effort and is not attempted. While extending the TC-Ontology, we mainly do the following: introduce new concepts/properties/axioms, modification in the existing constructs (e.g., splitting properties). In the following, we explain the extended definitions \& constructs.\\
\textit{a) Property split:} We analyze the existing properties and identify certain scenarios where splitting a property could enable us to perform more rigorous checks while validating the structures of various types of TC-AWP sentences. The sentence-structure validation helps identify the consistent and inconsistent cases. Intuitively, splitting the properties makes the system efficient, as the split enables the system to focus on a specific type of sentence at a given time. For example, existing TC-Ontology uses \textit{involvesSentence} object property to check the associations between a WP and its sentences of types BT, AT, and TR (refer Figure~\ref{propertysplitdetails}). With one of the property \textit{hasTR}, that is split from \textit{involvesSentence}, system can focus on TR-type sentences only. In contrast, we split the object property \textit{involvesAgent} (refer Figure~\ref{propertysplitdetails}) into three disjoint properties, which are: \textit{involvesAgentBT} with domain BT and range Agent, \textit{involvesAgentTR} with domain TR and range Agent, and \textit{involvesAgentAT} with domain AT and range Agent. In Figure~\ref{propertysplitdetails}, we mention the properties that we split. The \textit{involvesSentence} property is split based on the range type, whereas the \textit{involvesAgent} property is split based on the domain type.
% \begin{table}[h]
%     \centering
%     \begin{tabular}{|c|c|c|c|c|c|}
%     \hline
%         \textbf{Existing property} & \textbf{Domain} & \textbf{Range} & \textbf{Properties after split} & \textbf{Domain} & \textbf{Range}\\
%         \hline
%           &  &  & hasBT & Word-Problem & BT \\
%           involvesSentence  & Word-Problem & BT \ensuremath{\sqcup} AT \ensuremath{\sqcup} TR & hasAT & Word-Problem & AT \\
%           &  &  & hasTR & Word-Problem & TR \\
%         \hline
%           &  &  & involvesAgentBT & BT & Agent\\
%           involvesAgent  & BT \ensuremath{\sqcup} AT \ensuremath{\sqcup} TR & Agent & involvesAgentAT & AT & Agent\\
%              & &  & involvesAgentTR & TR & Agent\\
%          \hline
%     \end{tabular}
%     \caption{The table presents property split details}
%     \label{tablepropsplit}
% \end{table}
\\
\textit{b) Introducing new classes \& properties:} We model the following new classes to cater to the requirements in the current modeling: $has_{BT}$, $has_{AT}$, $has_{TR}$, $has_{QS}$, and $ValidStructure$. The details are discussed later in this section. To model the quantity involved in a \textit{transfer} and validate the structure of the TR-type sentences, we introduce a new object property \textit{involvesTR-Quantity} with domain TR and range TC-Quantity. This new property is required to generate the multiple transfer cases (explained in Section~\ref{sec:multitransfer}).
\\
\textit{c) Introducing new axioms to validate the structure of WP sentences and the word problem:} As mentioned earlier, the LSTM-generated TC-AWPs are categorized into three types- consistent TC-AWPs, partially consistent TC-AWPs  (repairable cases), and unrepairable cases. We propose new axioms to validate the structure of the sentences of generated cases. The BT and AT type sentences involve exactly one agent, and that agent owns exactly one TC quantity. We model the required knowledge about the structure of the BT and AT-type sentences using extended axioms EA.01 and EA.02 (given below). These axioms ensure the existence of information about an agent owning a TC quantity in BT and AT-type sentences. A TR-type sentence involves exactly one from-agent (one who loses the quantity), exactly one to-agent (one who gains the quantity), and exactly one transfer quantity. Therefore, by proposing the axiom EA.03, we aim to validate the structure of the TR-type sentences. Similarly, the axiom EA.04 helps validate the structure of the QS-type sentences. 
\\
\\
\textbf{Extended Axioms (EAs)}\\
EA.01: $Valid_{\small{BT}}$~\ensuremath{\equiv}~(= 1~\textit{involvesAgentBT.}(= 1 \textit{hasQuant.}~\textit{TC-Quantity})) ~\\
EA.02:  $Valid_{\small{AT}}$~\ensuremath{\equiv}~(= 1~\textit{involvesAgentAT.}(= 1 \textit{hasQuant.}~\textit{TC-Quantity})) ~\\
EA.03:  $Valid_{\small{TR}}$~\ensuremath{\equiv}~(= 1~\textit{fromAgent.}(= 1 \textit{hasQuant.}~\textit{TC-Quantity})~\ensuremath{\sqcap} = 1~\textit{toAgent.}(= 1 \textit{hasQuant.}~\textit{TC-Quantity})~\ensuremath{\sqcap} = 1 \textit{involvesTR-Quantity.}~\textit{TC-Quantity})~\\
EA.04:  $Valid_{\small{QS}}$~\ensuremath{\equiv}~(= 1~\textit{asksAbout.} \textit{Agent})~\ensuremath{\sqcap}~ ( = 1 \textit{asksObjType.} \textit{rdfs:Literal})
\begin{figure}[t]
    \centering
    \includegraphics[width=\textwidth]{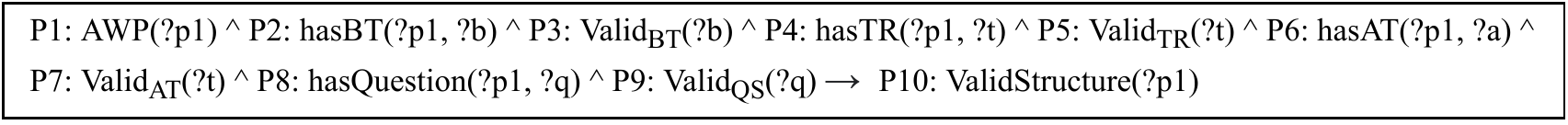}
    \caption{SWRL rule to validate the WP structure}
    \label{structureValid}
\end{figure}\\
\\
Our approach takes an LSTM-generated WP, extracts the concrete information from the problem text and populates it into the ontology ABox. The axioms devised (EA.01 - EA.04) processes the ABox information and verifies the validity of the sentence structures. For instance, axiom EA.01 validates the structure of a BT-type sentence and infers that the object representing BT-type sentence belongs to the $Valid_{BT}$ class. Proposed system makes use of the information inferred by these axioms in SWRL rule (given in Figure~\ref{structureValid}) and infers that the WP is generated with a valid structure. The explanation of the rule is as follows: In antecedent, predicates P1 and P2 verify the existence of BT type sentences in a word problem. Predicate P3 uses the information inferred by the axiom EA.01 to ensure that BT-type sentences are generated with a valid structure. Similarly, other predicates P4 - P9 uses the information inferred by the axioms EA.02 - EA.04 and ensure that other sentences are also generated  with a valid structure. Therefore, in consequent, predicate P10 makes the assertion representing that the WP at hand has a valid structure. Note that both consistent and partially-consistent TC-AWPs pass the structure validation test. We make use of this information in identifying consistent and repairing partially-consistent cases, explained in Sections~\ref{sec:identifyConsistentWPs} \&~\ref{sec:repairingWPs}, respectively. Proposed system rejects LSTM-generated WPs which do not pass the structure validation test, and they are not included in the successfully generated cases.
%\vspace{-1mm}
\subsection{ABox Extraction - Sentence Classification and Populating the Ontology}
\label{sec:abox}
\textit{a) Sentence Classification}:
As mentioned in the introduction section, the TC-AWP domain consists of four types of sentences: before-transfer (BT), transfer (TR), after-transfer (AT), and question (QS). The sentence classification model aims to assign a type to each WP sentence based on its content. The information about the type of a sentence is helpful at various stages of the system, for example- information extraction, inconsistency detection, etc. Also, the sentence-type information helps obtain a more meaningful description of the problem text and capture important information such as the direction of a transfer, agent, and quantity about which posed question seeks an answer. We normalize the agent names present in the problem-text using NLTK \citep{nltk} python library (by replacing the actual names with Agent1, Agent2, etc.) to improve the sentence classification accuracy. During feature engineering, we extract unigrams, bigrams and trigrams features in addition to the Bag of Words (BoW) features. Also, we use the positions of the sentences as an additional feature. For example, BT-type sentences appear at the beginning and QS-type sentences towards the end in a problem-text. We experiment with boosting models \citep{xgboost} (AdaBoost and XGBoost) and BERT-based model \citep{bert} for the sentence classification task. Features discussed above are used when leveraging Boosting models. Since the BERT-based classifier achieves superior results and is current state-of-the-art, we deploy BERT for the sentence classification task. Related results are provided in Section~\ref{sec:experiment}.
\begin{figure}[t]
    \centering
    \includegraphics[width=0.9\textwidth]{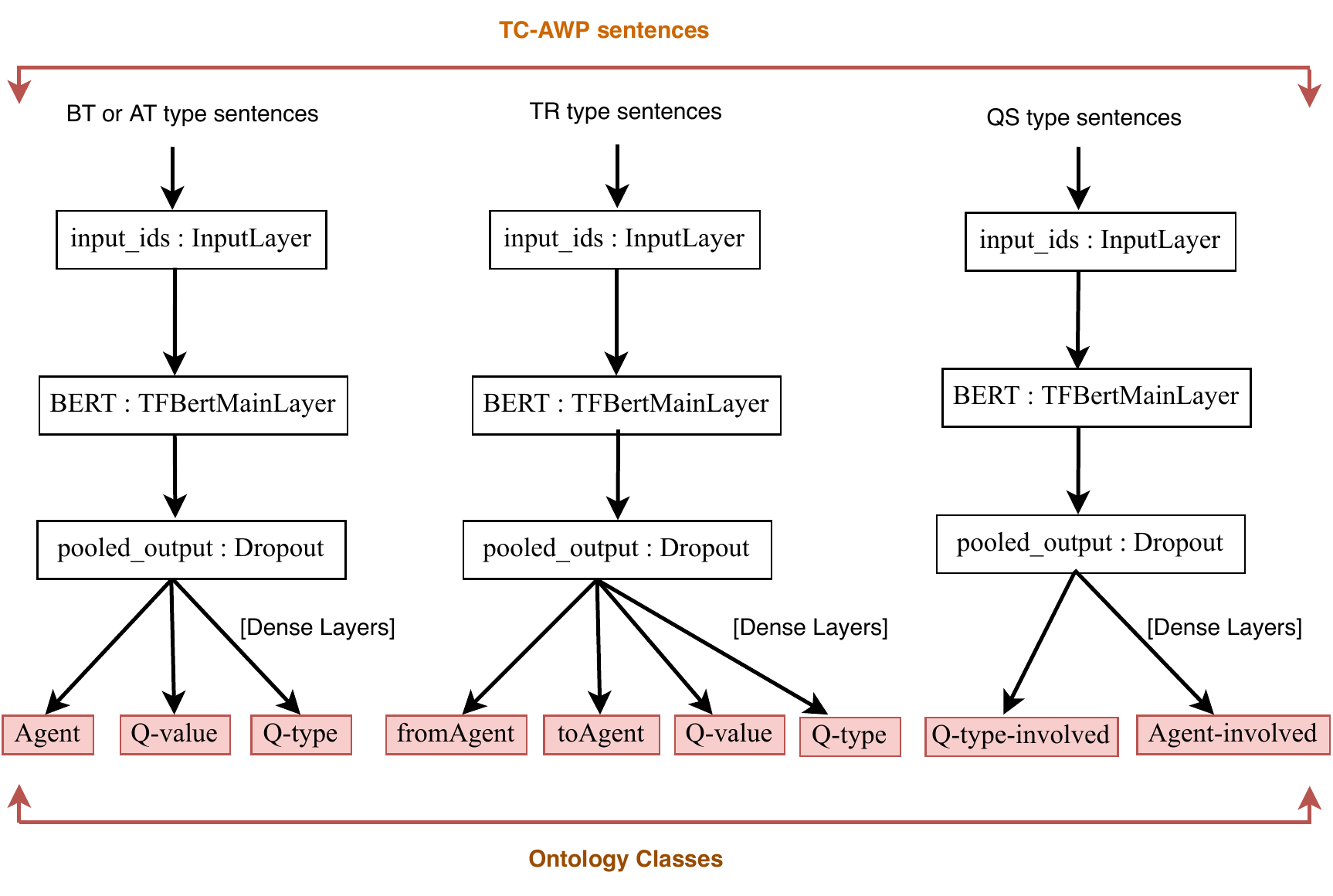}
    \caption{BERT architecture}
    \label{fig12}
\end{figure}\\
\textit{b) Populating the Ontology with ABox information}: The ABox information that populates into the ontology differs based on the sentence type. The sentence classification module provides sentence-type information to the system. While developing the domain ontology, we model the classes and properties in such a way that the information extracted from TC-AWP sentences can be appropriately populated into the ontology. For example, a BT-type sentence such as "Agent1 has 5.0 books" contributes the following ABox assertions: "Agent1" is an individual of \textit{Agent} class, quantity Q1 is an individual of \textit{TC-Quantity} class, and object property assertion (<tc:Agent1 tc:hasQuant tc:Q1>) and related data property assertions. BERT-based LM \citep{bert}, designed to pre-train bidirectional representations from the unlabeled text, is utilised by us to automatically populate the ontology ABox.
Bidirectional training helps BERT learn a more profound sense of language context. Since it is possible to \textit{fine-tune} the pre-trained BERT model on a domain of interest and BERT is empirically powerful in modeling downstream tasks such as question answering, language inference, etc., we choose BERT as a LM.
\begin{figure}[t]
    \centering
    \includegraphics[width=0.9\textwidth]{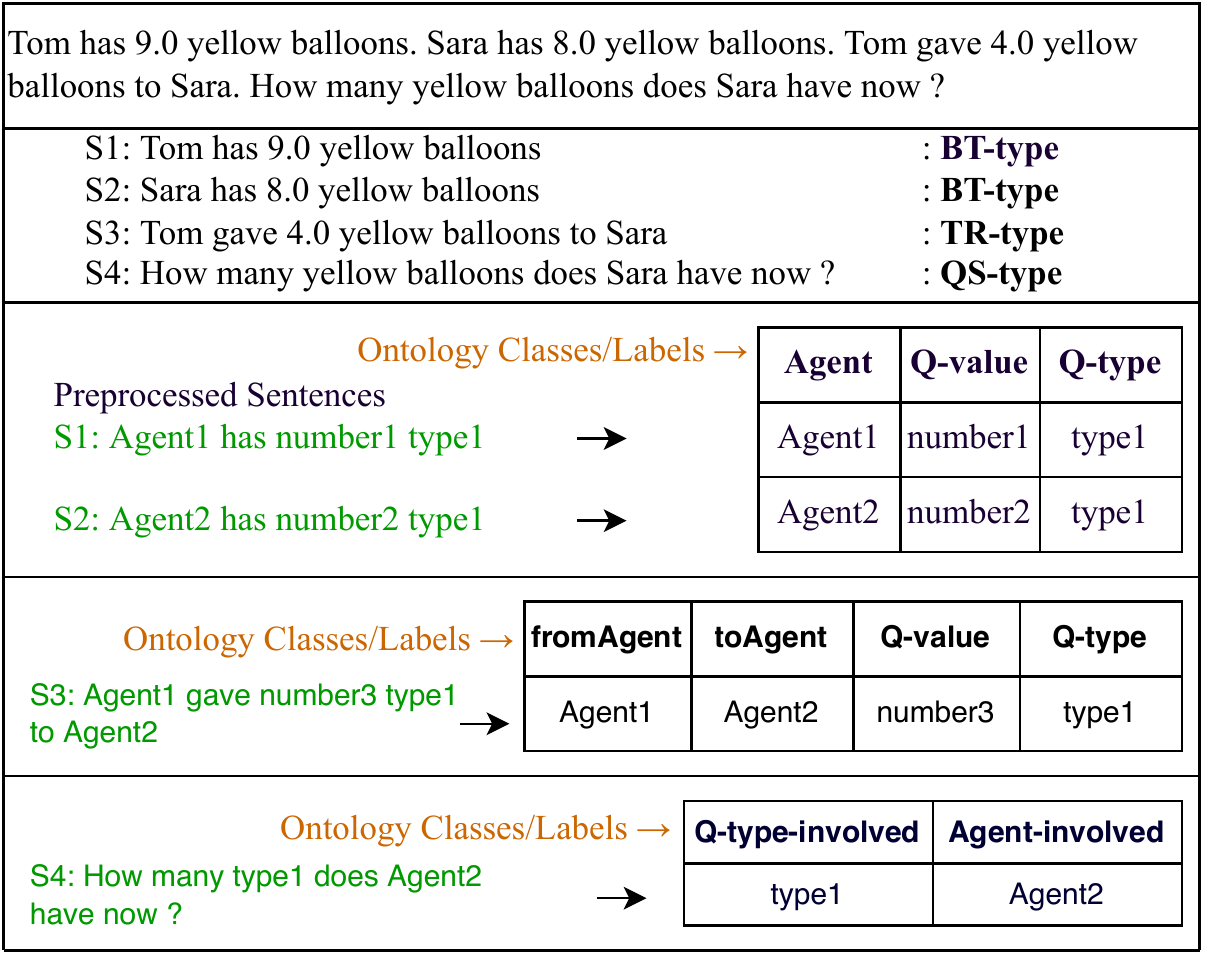}
    \caption{Ontology classes and sentence-parts that become the class instances}
    \label{fig11}
\end{figure}\\
As shown in Figure~\ref{fig11}, we preprocess the WP sentences to replace agent-names by AgentX, actual numbers by numberY, etc. Sentence preprocessing is required to apply the BERT LM for ABox extraction, as the probable token values (e.g., AgentX, numberY, etc.) for a sentence-part should be finite. However, we keep the information about mapping the original value to its replaced string. These replacements are performed using POS tag values which we get using NLTK \cite{nltk} Python library. Sentence-parts carry information on class-instances of the ontology. For example, in the preprocessed S1 sentence given in the Figure~\ref{fig11}, "Agent1" is an instance of the class \textit{Agent}. Therefore, we annotate the sentence parts with class names from TC-Ontology and train a BERT LM that learns to produce instances for these ontology classes from the given WP sentences. In Figure~\ref{fig11}, we present an example word problem and show various sentence parts that contribute to the ontology ABox. The given TC-AWP is composed of four sentences where S1 and S2 are BT type, S3 is TR type, and S4 is QS type. Since the structure of these various types of sentences differs, the information they carry also differs. Therefore, the ontology classes required to hold the instance information differ based on the type of sentence. For example, in Figure~\ref{fig11}, the ontology classes mentioned against BT-type sentences (S1 and S2), the TR-type sentence (S3), and QS-type sentence (S4) are different.\\
\textit{Fine-tuning the language model:} The fine-tuning approach introduces minimal downstream-task-specific parameters and fine-tunes all the pre-trained parameters. In Figure~\ref{fig12}, we present the architecture of the language model to explain the fine-tuning required in our system. As shown in the Figures~\ref{fig11} and~\ref{fig12}, in our case, the downstream task is predicting the instances for ontology classes.
Since the sentences of TC-AWPs differ in structure and information, we use a separate BERT language model for each sentence category. However, the information carried by the BT and AT types of sentences belongs to the same ontology classes. Therefore the same LM can be used to learn information from these two types of sentences. We choose \textit{bert-base-uncased} as a pre-trained model. Since we follow the fine-tuning procedure similar to the one given in \citep{bert} (except that we propose using ontology classes as labels) and train the parameters as explained in \citep{bert}, we omit the detailed background description of the language model architecture. We discuss the results related to the task in Section~\ref{sec:experiment}.
\subsection{Identifying Consistent TC-AWPs}
\label{sec:identifyConsistentWPs}
\begin{figure}[t]
    \centering
    \includegraphics[width=\textwidth]{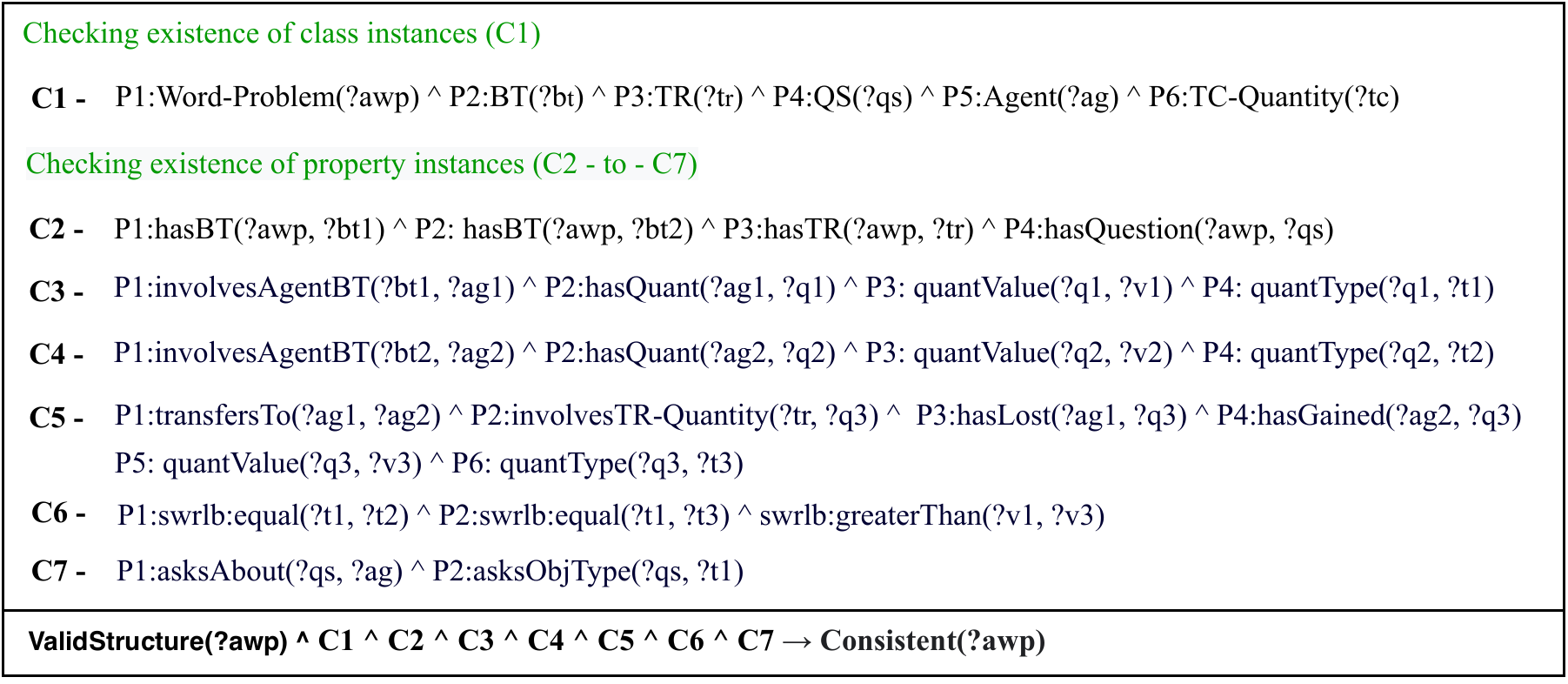}
    \caption{Example SWRL rule showing how to identify a consistent TC-AWP (\textit{explanation} is given in the text)}
    \label{consistentWP}
\end{figure}

Automatically identifying the consistent word problems, from the LSTM-generated output, is important, as it  eliminates the human intervention. We analyze the major causes that can make TC-AWPs inconsistent and incorporate the domain knowledge required to check the consistency of the TC-AWPs. The various factors that can affect the consistency of TC-AWPs are: math operation, question posed, missing information, etc. For instance, addition and subtraction operations due to an object transfer are dependent on several other factors such as: the subtraction operation should produce non-negative number as we deal with the transfer WPs, compatibility of object types, etc. Also, sometimes object-type element in the question-sentence is generated wrongly (for example - question queries about object-type "red marble", and other sentences talk about object-type "red balloon"), which causes inconsistency due to the question posed. By missing information, we mean that system should check whether important information such as agent-elements, object-type-elements, transfer-quantity, etc. are present in the generated word problem. Since ABox information (class and property instances) is available inside the ontology, SWRL rules (we develop and made them available inside the ontology itself) can appropriately check the consistency of the word problems.\\
\\
Consider an example TC-AWP: "Stephen grew 72 carrots. Daniel grew 25 carrots. Stephen gave 13 carrots to
Daniel. How many carrots does Daniel have now?".  In Figure~\ref{consistentWP}, we discuss SWRL rule that shows how the proposed system checks the consistency of the given word problem. The \textit{explanation} of the same is as follows: C1 component of the rule ensures that the generated word problem consists of all the required class instances. C2 component checks the property instances and identifies the presence of two BT type, one TR type and one QS type sentences. C3 and C4 components check that both the BT type sentence are generated with agent and quantity information. C5 component verifies that the TR type sentence is generated appropriately. C6 and C7 components verify that the object type elements in all the sentences are generated appropriately and object transfer is possible (as agent1 is transferring less than from what he owns). Given that all the predicates mentioned in C1 -to- C7 components are true and WP has a valid structure, system considers the given TC-AWP consistent. The graded style of consistency checking could enable the system to identify the appropriate portion of the word problem that is generated wrongly. TC-AWPs essentially involve BT, TR, and QS types sentences. The AT type sentences are optional. Word problem discussed above does not involve an AT type sentence. The consistency of the TC-AWPs that involves AT type sentences can be checked in similar manner.
\subsection{Identifying \& Repairing Partially Consistent TC-AWPs}
\begin{figure}[!h]
    \centering
    \includegraphics[width=\textwidth]{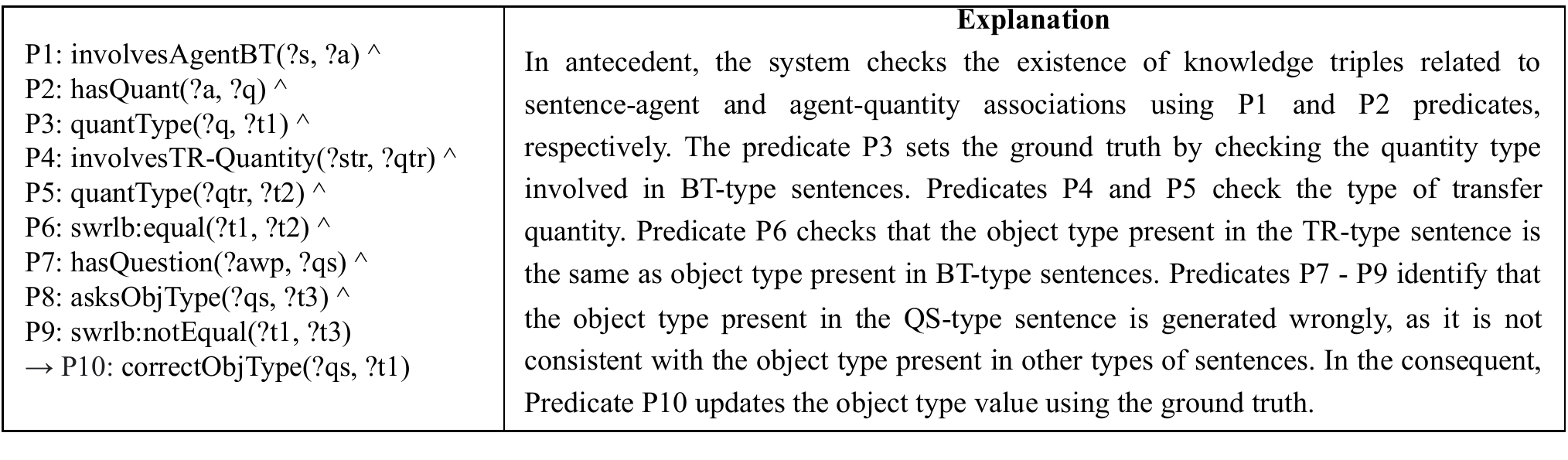}
    \caption{SWRL rule to fix object type element in QS-type sentence}
    \label{repairQS}
\end{figure}

\begin{figure}[t]
    \centering
    \includegraphics[width=\textwidth]{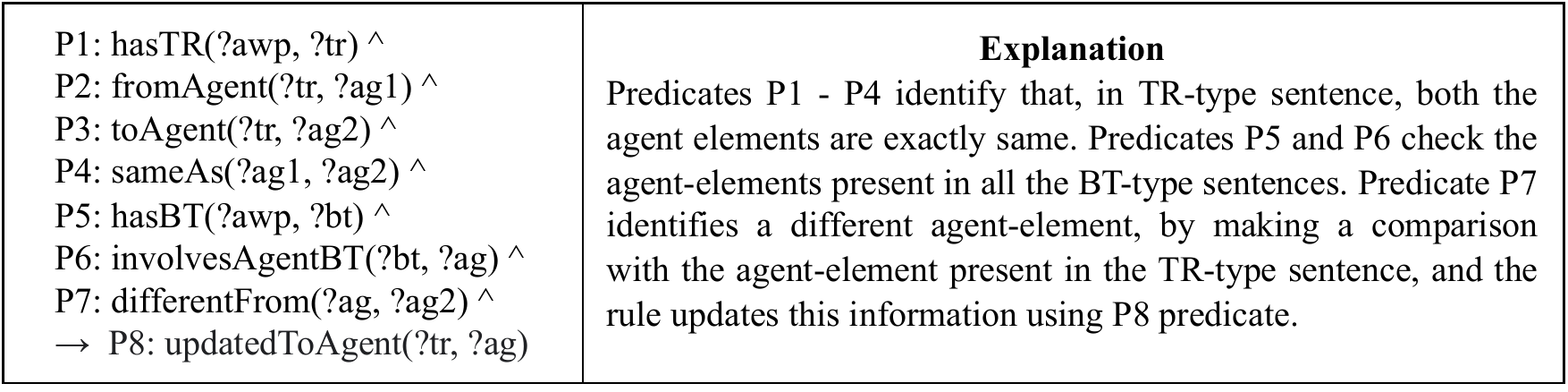}
    \caption{SWRL rule to fix agent element in TR-type sentence}
    \label{repairTR}
\end{figure}

\begin{figure}[t]
    \centering
    \includegraphics[width=\textwidth]{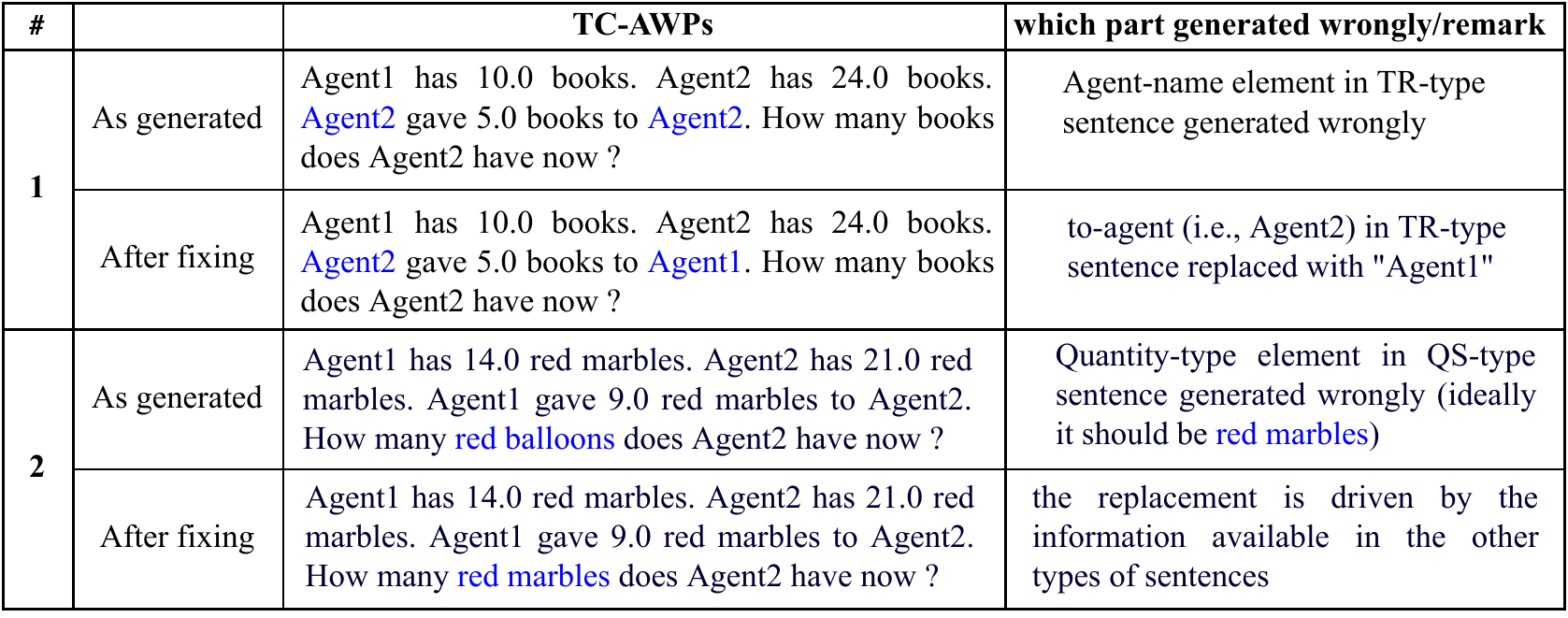}
    \caption{Analysis of Ontology based post processing}
    \label{ResPostProcess}
\end{figure}

\label{sec:repairingWPs}
As mentioned earlier, though we train the LSTM model on consistent TC-AWPs, it generates partially-consistent examples too. This section explains how the proposed system repairs partially-consistent-cases and makes them consistent. Note that partially-consistent-cases are those TC-AWPs in which one or two elements are generated wrongly. The domain knowledge that helps in the repairing process is encoded using ontology TBox and SWRL rules. BERT-based LM (explained in Section~\ref{sec:abox}) helps the system extract ABox information from the problem text. Therefore, the consolidated knowledge (i.e., TBox and ABox) is available in the ontology. The inconsistency detection module (part of check \& repair module, see Figure~\ref{fig4}) processes the consolidated knowledge and identifies the category of the generated TC-AWP text. For example, in Figure~\ref{fig4}, AWP1, AWP2, and AWP3 are labeled as consistent, partially consistent, and unrepairable, respectively. The partially consistent cases are fetched to the repair module. 
In the following we discuss some inconsistent examples and the SWRL rules that repair these examples. Note that the rules discussed in Figures~\ref{repairQS} \&~\ref{repairTR} use the predicate \textit{ValidStructure(?awp)} to ensure that the WP has a valid structure, however, its not included in the rules.\\
\textit{a) AWP examples where object-type element in the QS-type sentence is generated wrongly:} Under this category, the system considers all such LSTM-generated examples that are inconsistent due to the wrongly generated object-type element in the QS-type sentence. However, repairing of such examples requires setting a ground truth. To address the issue, we consider the object-type present in the BT-type sentences as the ground truth. Consider the following example: "\textit{Agent1 has 5.0 red balloons. Agent2 has 4.0 red balloons. Agent1 gave 2.0 red balloons to Agent2. How many red marbles does Agent1 have now?}", where object-type in QS-type sentence is generated wrongly. The SWRL rule explained in Figure~\ref{repairQS} repairs the word problem discussed above.
\\
\textit{b) AWP examples where agent elements involved in a transfer are generated wrongly:} Under this category, the system considers all such LSTM-generated TC-AWP texts inconsistent due to wrongly generated agent elements in transfer-type sentences. Consider the following example: "\textit{Agent1 has 15.0 pens. Agent2 has 12.0 pens. Agent2 gave 2.0 pens to Agent2. How many pens does Agent1 have now?}", where agent element in TR-type sentence is generated wrongly. The SWRL rule explained in Figure~\ref{repairTR} repairs the WP mentioned above.
\\
\textit{c) Other possible partially consistent cases:} As explained in Section~\ref{sec:extconto}, the LSTM model always generates BT-type sentences correctly. The TR, AT, and QS type sentences, make the generated WP partially consistent mainly due to the wrongly generated elements in these sentences, which possibly are: object-type, agent, and transfer-quantity (considered wrongly generated when the amount of quantity involved in a \textit{transfer} exceeds the amount an agent owns), etc. In Figures ~\ref{repairQS} \& \ref{repairTR}, we show how the proposed system repairs TC-AWPs where object-type or agent-elements were generated wrongly. The other partial consistent cases where AT and TR-type sentences generated with wrong object type are repaired in a similar manner.\\
In Figure~\ref{ResPostProcess}, we present two examples which are made consistent by the "Check \& Repair" module. In example 1, we show that both the agent-elements present in the TR-type sentence are same, which makes the given WP partially consistent. The encoded domain knowledge helps the system in identifying the inconsistency and makes the WP consistent by replacing to-agent (i.e., Agent2) by "Agent1".
% \begin{table}
%     \centering
%     \begin{tabular}{|c|c|>\justify m{8cm}|m{4.8cm}|}
%     \hline
%   \# &   & \hspace{3cm}\textbf{TC-AWPs}& \textbf{which part generated wrongly/remark}\\
%  \hline
%   1  & As generated & Agent1 has 10.0 books. Agent2 has 24.0 books. \textcolor{blue}{Agent2} gave 5.0 books to \textcolor{blue}{Agent2}. How many books does Agent2 have now ? & Agent-name element in TR-type sentence generated wrongly\\
%     \hdashline
%   & after fixing & Agent1 has 10.0 books. Agent2 has 24.0 books. \textcolor{blue}{Agent2} gave 5.0 books to \textcolor{blue}{Agent1}. How many books does Agent2 have now ? & to-agent (i.e., Agent2) in TR-type sentence replaced with ``Agent1''\\
%  \hline
 
%   2  & As generated& Agent1 has 14.0 red marbles. Agent2 has 21.0 red marbles. Agent1 gave 9.0 red marbles to Agent2. How many \textcolor{blue}{red balloons} does Agent2 have now ? & Quantity-type element in QS-type sentence generated wrongly\\
%     \hdashline
%   & after fixing &  Agent1 has 14.0 red marbles. Agent2 has 21.0 red marbles. Agent1 gave 9.0 red marbles to Agent2. How many \textcolor{blue}{red marbles} does Agent2 have now ? &  the replacement is driven by the information available in the other types of sentences\\
   
%   \hline
% \end{tabular}
%     \caption{Analysis of Ontology based post processing}
%     \label{postprocessing}
% \end{table}

\section{Generating multiple transfer cases}
\label{sec:multitransfer}
\begin{figure}
    \centering
    \includegraphics[width=\textwidth]{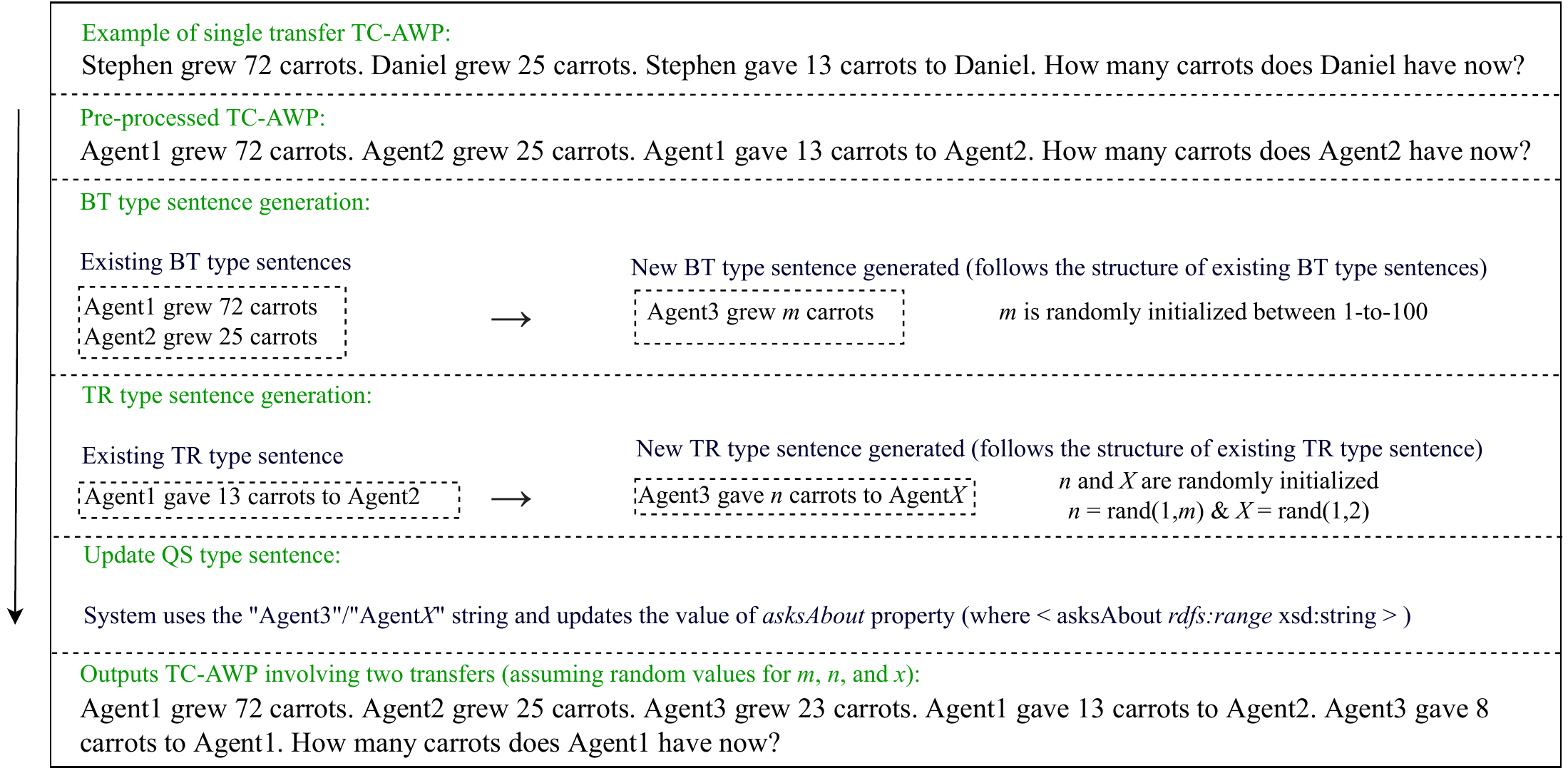}
    \caption{Illustration showing generation for multi transfer TC-AWP from single transfer TC-AWP}
    \label{multiTransferGeneration}
\end{figure}

\begin{figure}[!h]
    \centering
    \includegraphics[width=0.8\textwidth]{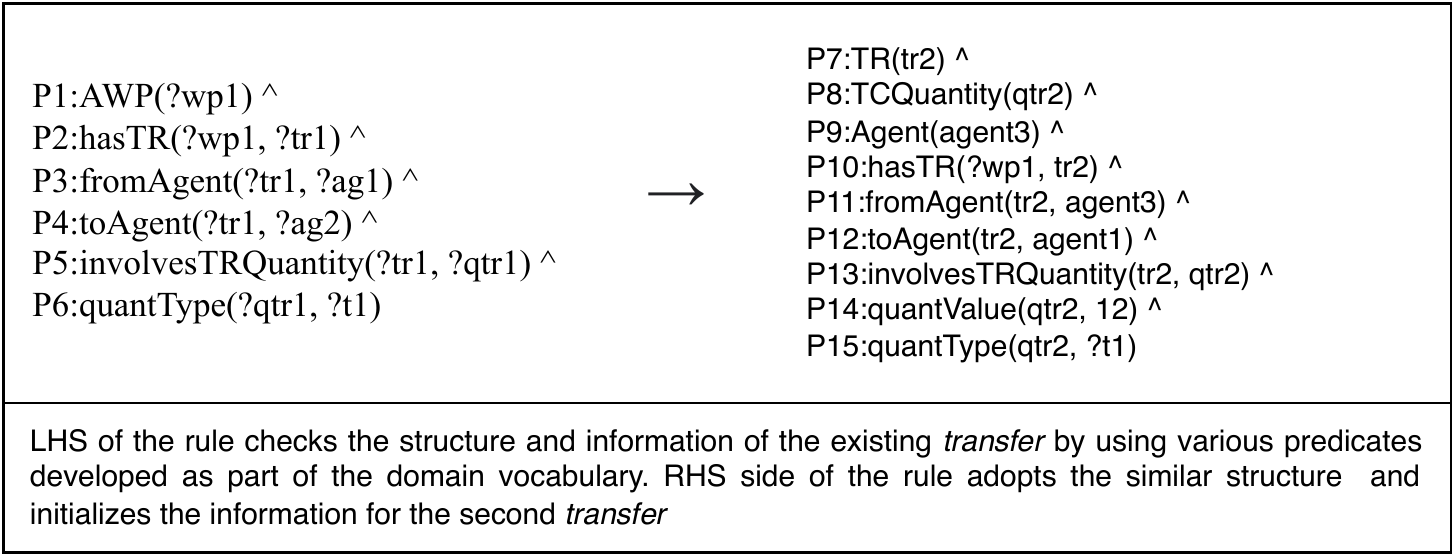}
    \caption{SWRL rule showing the generation of second transfer}
    \label{secondTransfer}
\end{figure}

Existing AWP datasets contain TC word problems that involve only one object transfer. For example, the TC-AWP given in Figure~\ref{multiTransferGeneration} involves one object transfer, i.e., "Stephen gave 13 carrots to Daniel".
If a word problem involves more than one object transfer, we call it a multiple transfer case (or multi-transfer TC-AWP). The existing word problem solvers may find it challenging to solve multi-transfer TC-AWPs, as the subsequent transfer depends on the previous transfer, and sequential reasoning is needed. We focus on generating multi-transfer TC-AWPs, as they could be useful in training a robust solver system and also tests the capability of the NLU component of the solver system. As we can see from the results mentioned in Table~\ref{tab4}, the probability of generating consistent TC-AWP using SOTA Deep learning methods drops drastically when the number of sentences in TC-AWPs increases. Therefore, intuitively, the probability of generating consistent multi-transfer TC-AWPs by SOTA Deep learning methods will be quite low, as the number of sentences in TC-AWPs will increase (a typical two-transfer TC-AWP will involve 7 sentences). Our ontology-based approach can ensure that every generated TC-AWP is consistent. Next, we discuss our approach for multi-transfer TC-AWP generation.\\
To generate a meaningful multi-transfer TC-AWP from a single-transfer TC-AWP, we devise the following strategy: generate an additional BT-type sentence involving third agent, a TR-type sentence with \textit{transfer} happening between third agent and one of the existing agents, and update the QS-type sentence accordingly. In Figure~\ref{multiTransferGeneration}, we present an illustration to show how a two-transfer WP is generated from a single-transfer WP.
\\
\textit{Generation process:} First, we randomly initialize the additional ABox information, i.e., instances of the ontology classes such as agent, TC-Quantity, TR, etc. The proposed system uses this extra piece of ABox information and the structural information obtained from the triples of the first transfer and generates the second transfer. In Figure~\ref{secondTransfer}, we show the SWRL rule, where the antecedent part traverses through the triples of the first transfer, and the consequent part generates a similar structure and information for the second transfer. The natural language interpretation of the LHS of the rule in plain English is as follows: "The given AWP-text includes a TR-type sentence involving from-agent (one who loses the quantity), to-agent (one who gains the quantity), transfer quantity, and type of the quantity". The triples pertaining to an additional BT-type sentence can be generated either using a program script or an SWRL rule; however, we use an SWRL rule. The process of generating an additional BT-type sentence and updating the QS-type sentence is simple; therefore, we skip the discussion. Since the underlying ontology (which have the required domain knowledge to check the consistency of TC-AWPs) drives the generation process, the generated multi-transfer cases are consistent.
\\
All the sentences (natural language text) of the single-transfer WP are available in the ontology. However, the newly generated sentences (one BT and one TR type) are available in the form of triples only. Therefore, we use these triples and perform template based conversion and obtain a plain English text for these new sentences. An example of the generated multi-transfer TC-AWP is shown in the Figure~\ref{multiTransferGeneration}.
Note that there are multiple ways in which a second transfer can be additionally included in a single-transfer AWP. The second transfer could be from Agent1 to Agent3 or Agent2 to Agent3 or from Agent3 to either of Agent1 or Agent2. In the above discussion we have given the details of the last case only. Also, the question can be about any of the three Agents. Thus we get about 12 combinations. We skip the discussion on other cases as it is easy to see that cases are similar and the techniques we have provided above can be used to implement the cases.
\section{Implementation Details and Analysis of Experiments}
\label{sec:experiment}
While implementing the proposed system, the following sets of tasks were performed: LSTM-based TC-AWP generation, modeling domain knowledge (TBox design and developing SWRL rules) by developing an ontology in Prot\'eg\'e tool\footnote{\url{https://protege.stanford.edu/}} \citep{protegetool}, and the ABox extraction using BERT LM.
For ontology modeling, we installed the Prot\'eg\'e tool on a system with an Intel Core-i7 processor,  32 GB RAM, and Ubuntu 18.04 OS. For LSTM (word problem generation) and BERT (sentence classification and ABox extraction) related experiments, we chose Python as a programming language and used Google-Colab\footnote{
https://colab.research.google.com}. To access the domain ontology in Google-Colab, we use the Owlready2\footnote{https://owlready2.readthedocs.io/en/v0.36/} \citep{owlready2} Python package, which allows ontology-oriented programming.
Since we are the first to propose an approach (to the best of our knowledge) that focuses on checking the consistency of the generated word problems, the state-of-the-art systems are not available and we could not provide a comparative analysis of results. However, we provide a detailed analysis on the results of the system's sub-tasks such as LSTM-based TC-AWP generation, sentence classification, and class-instance prediction using BERT LM; and provide examples that were repaired by leveraging the domain knowledge.
\subsection{Details of the datasets used}
\label{sec:expdatasets}
In the introduction section, we briefly discussed the popular AWP datasets. In the current work, we primarily focus on AllArith, ALG514 , MAWPS, and Dolphin-S datasets, as they contain TC-AWPs expressed in the English language. Since our domain of study is transfer case AWPs, we gather TC-AWPs from these above mentioned datasets by using a keyword-based program script and name the dataset Arith-Tr. Arith-Tr consists of 694 TC-AWPs. The LSTM-based generator model is trained using Arith-Tr.\\
% To generate TC-AWPs, we make use of Arith-Tr dataset to train the LSTM-based generator model.\\
To prepare the training data for sentence classifier, we take the TC-AWPs from Arith-Tr and split them into various types of sentences. We label these sentences manually and call this labeled dataset "TC-Sentences". The details on the datasets used for training LM for predicting the instances of ontology classes are provided in Section~\ref{sec:bertexp}.
\subsection{Results of LSTM-based TC-AWP generation}
% In Figure~\ref{fig3}, we presented the architecture of LSTM-based TC-AWP generation model. 
As mentioned earlier, the LSTM model generates TC-AWPs of three kinds. In Table~\ref{tab4}, we mention the proportion of these three kinds of TC-AWPs in the generated cases. The first row indicates that all the WPs in the training set include three sentences and the LSTM model generates WPs of similar length. The results show that as we increase the length of the word problem text, the chances of generating consistent WPs decrease drastically. Results also highlight that a large number of cases (40\% when WPs include five sentences) are generated with minor errors (only one or two elements are generated wrongly). Table~\ref{table1} presents some examples of LSTM-generated TC-AWPs.
\begin{table}
    \centering
    \begin{tabular}{|l|c|c|c|c|}
    \hline
        \textbf{Model}&\# \textbf{Sentences} & \textbf{\% of Consistent}&\textbf{\% of Repairable}&\textbf{\% of Unrepairable} \\
        &\textbf{in WPs} & \textbf{ TC-AWPs}&\textbf{Cases}&\textbf{Cases} \\
        \hline
        & 3 & 94 & 4 & 2\\
        LSTM  & 4 & 53 & 28 & 19\\
         & 5 & 36 & 40 & 24\\
         \hline
    \end{tabular}
    \caption{Proportion of different kinds of TC-AWPs in LSTM generated examples (with varying number of sentences)}
    \label{tab4}
\end{table}

\begin{table}
    \centering
    \begin{tabular}{|c|>\justify m{8.1cm}|c|}
    \hline
  \# & \centering \textbf{Generated AWPs} & \textbf{Validity}\\
 \hline
 1&  Agent1 had 74 dollars. Agent1 donated 6 dollars to Agent2. How many dollars does Agent1 have now ? & Consistent \\
  \hdashline
%   2&  Agent1 has 7.0 apples. Agent2 has 11.0 apples. Agent2 gave Agent1 8.0 apples. How many apples does Agent1 have now ? & Consistent \\
%   \hdashline
  2&Agent1 has 44.0 blue marbles. Agent2 has 24.0 blue marbles. Agent2 gave 13.0 blue marbles to Agent1. How many blue balloons does Agent1 have now ? & \textcolor{blue}{Partially Consistent (QS)}\\
    \hdashline
    %  4& Agent1 has 6.0 pens. Agent2 has 7.0 pens. Agent2 gave 4.0 pens to Agent2. How many pens does Agent1 have now ?& \textcolor{blue}{Partially Consistent (TR)}\\
    % \hdashline
      3& Agent1 has 9 pens. Agent1 gave Agent2 gave 4 pens to Agent1. How many pens does Agent1 have now ? & \textcolor{red}{Unrepairable}\\
    %   \hdashline
    %   6& Agent1 had 17.0 marbles. Agent2 had 34.0 blue plates. Agent1 now has 22.0 seashells. How many seashells does Agent2 have now ?& \textcolor{red}{Unrepairable}\\
      \hline
      \end{tabular}
    \caption{Examples of LSTM-generated TC-AWPs. \textcolor{blue}{Partially Consistent (QS)} represents that QS-type sentence is making the WP partially consistent.}
    \label{table1}
\end{table}
%\vspace{-10mm}
\subsection{Sentence Classification}
We use "TC-sentences" dataset (explained in Section~\ref{sec:expdatasets}) and apply the standard train-test split for assessing the sentence classification task. Initially, we used a Python open-source scikit-learn library to deploy the boosting models (AdaBoost and XGBoost) and found that both the models achieved 95\% accuracy on the test data. We analyze the results and found that boosting models predict TR and QS-type sentences with 100\% accuracy, as their structure is different from other types of sentences and also they consist of some unique words. Whereas, the miss-classification rate (5\%) is mainly due to the BT and AT types of sentences, as sometimes these two types of sentences are very similar. For example- Stephen has 5 pens \textit{and} Stephen now has 5 pens. The first sentence is of BT-type whereas the later one is of AT-type, while they differ in only one word.
Later, we fine-tune a pre-trained BERT LM using "TC-Sentences" dataset. While fine-tuning, we use ADAM optimizer and keep the learning rate $5e^{-5}$. BERT-based classifier achieves 100\% accuracy on the test data for the sentence classification task and we adopt it in our solution.
\subsection{BERT language model for prediction of instances of ontology classes}
\label{sec:bertexp}

\begin{table}
  \centering
    \begin{tabular}{|c|l|c|c|}
    \hline
    \textbf{Sentence Type} & \textbf{Ontology Classes} & \textbf{1K Sentences}$^{*}$ & \textbf{2K Sentences}$^{*}$\\
    \hline
    & Agent & 95.5 & 98.0\\
  BT/AT  & Q-Value & 93.2 & 95.4 \\
      & Q-Type & 95.0 & 97.2\\
      \hline
  \multirow{4}{*}{TR}    & From-Agent & 83.4 & 94.3 \\
      & To-Agent & 83.0 & 95.0\\
        & Q-Value & 78.6 & 87.2 \\
         & Q-Type & 80.1 & 88.5 \\
         \hline
     \multirow{2}{*}{QS}     & Agent & 95.2 & 98.7\\
          & Q-Type & 88.3 & 95.2\\
          \hline
          \multicolumn{2}{|c|}{Joint Accuracy (\%)} & 64.0 & 74.6 \\
      
     \hline
    \end{tabular}
    \caption{Assessment of the BERT LM for class-instance prediction task (accuracies are on \% scale). $^{*}$ refers to more details about the datasets which are available in Section 6}
  	\label{table:theLM}
\end{table}
In this sub-section, the word "class" is to be understood as the Classes in the TC-Ontology. 
For the prediction of instances of ontology classes, we manually annotate the WP sentence-parts with the class-names from the TC-Ontology and prepare the labeled-datasets (namely AB-Sentences, TR-Sentences, QS-Sentences), as shown in Figure~\ref{fig11}, for the BERT LM. The LM needs three different datasets (mentioned above) as each sentence category contributes a separate piece of information; therefore, each dataset's labels (i.e., ontology classes) are entirely different. Since BT and AT-type sentences have a similar structure and labels are the same, we create only one dataset (i.e., AB-Sentences). As shown in Table~\ref{table:theLM}, we train the LM on 1K and 2K sentences, where proportion of various types of sentences is: AB-Sentences (53.55\%), TR-Sentences (23.22\%), and QS-Sentences (23.22\%). We adopt a stratified 5-fold cross-validation procedure to randomly allocate the data-points to the training and testing data splits. As shown in Figure~\ref{fig12}, we make three instances of the pre-trained LM and train each model on different type of sentences (BT/AT or TR or QS-type). Table~\ref{table:theLM} presents the results of these three LMs. When we consider 2K sentences for assessment, all the LMs achieve high accuracies ($\approx 95\%$) while predicting instances for most of the ontology classes. However, the LMs achieve 74.6\% joint accuracy (refers to percentage of WPs for which all the class instances are predicted correctly). 
\subsection{Analysis of results}
Since BERT-based LM achieves 74.6\% joint accuracy in predicting the instances of ontology classes, it limits the ability of the repair process of the proposed system. As shown in Table~\ref{tab4}, the LSTM model generates 40\% partially consistent cases; however, the system was able to repair only 74.6\% of these cases. Moreover, when we manually analyzed the samples of the consistent TC-AWPs, we observed that our system wrongly considered some cases consistent. Fortunately, the miss-classification rate is less than 1\%. For example: "\textit{Agent1 has 24.0 books. Agent2 has 12.0 books. Agent2 took 13.0 books to Agent1. How many books does Agent1 have now ?"}. In this example the TR-type sentence should have been "Agent2 took 13.0 books from Agent1". In the current modeling, we primarily focus on checking the mathematical consistency of the generated WPs, handling the consistency in natural language variations is out of the scope of this work.
\subsection{Single-transfer and multi-transfer TC-AWPs generated by the proposed approach}
A sample of the generated TC-AWPs is made available on google drive (\href{https://drive.google.com/file/d/1fdQulVEVVvhDp-I59x7oBwDxSKbm8Xhx/view?usp=sharing}{link}).

\section{Conclusion \& Future Directions}
\label{sec:conclusion}
% For the task of word problem generation, we show that even though deep neural networks are trained on the consistent word problem text, they generate inconsistent problems too. Human intervention to analyze and identify the consistent cases from the generated ones is time-consuming and may even be error-prone. Therefore, there is a need for an automatic consistency checker. We believe this can only be achieved using ontology-based methods, as appropriate domain knowledge can be encoded and incorporated. Therefore, we demonstrate the use of domain ontology in checking the consistency of the generated word problems. Also, we identify interesting logical situations where domain knowledge can help identify \& fix the inconsistencies. We model domain knowledge (related to TC-AWPs) using OWL-DL ontology (named as TC-Ontology) and show the use-case in converting the partially-consistent-cases into the consistent ones. Further, we show the use of domain ontology in generating multi-transfer TC-AWPs from single-transfer TC-AWPs. We feel that the following are the interesting research directions to explore: a) a more efficient way to extract the ontology ABox and b) include domain knowledge that can help check the inconsistency due to the natural language variations.
The work investigated the challenges in machine generation of mathematical texts. Purely neural-based generation lacks consistency whereas ontology-based generation is deficient in producing natural language variations. Also, human intervention to analyze the machine-generated output to identify the consistent cases is time-consuming and may even be error-prone. A combined approach where an LSTM-based module generates reasonably good quality mathematical text and an ontology-based module (by exploiting domain knowledge) provides a facility to check the consistency of the LSTM-generated output has turned out to be effective. As a bonus, the ontology-based module is also able to repair a portion of the generated problems as these cases are partially consistent.  The reason why consistency checking and repairing can be effectively achieved using ontology-based methods is that appropriate domain knowledge can be encoded and exploited. Further, we also demonstrated the use of the same domain ontology in generating multi-transfer 3-agent TC-AWPs from single-transfer TC-AWPs. The interesting future research directions to explore are: a) a more efficient way to extract the ontology ABox, b) include domain knowledge that can help check the inconsistency that arises due to the natural language variations, and c) adopting the proposed approach to generate other types of AWPs.
% \appendix
% \input{sections/appendix}
% \begin{acknowledgments}
% We thank the reviewers and action editor for their thoughtful comments and suggestions.
% This work was partially supported by the Defense Advanced Research Projects Agency (DARPA) under the World Modelers program, grant \#W911NF1810014, and by the National Science Foundation (NSF) under grant \#2006583. 
% Mihai Surdeanu declares a financial interest in \url{lum.ai}. This interest has been properly disclosed to the University of Arizona Institutional Review Committee and is managed in accordance with its conflict of interest policies.
% \end{acknowledgments}

\starttwocolumn
\bibliography{main}
\clearpage
\end{document}